\def\year{2022}\relax
\documentclass[letterpaper]{article} 
\usepackage{aaai24}  
\usepackage{times}  
\usepackage{helvet}  
\usepackage{courier}  
\usepackage[hyphens]{url}  
\usepackage{graphicx} 
\usepackage{natbib}  
\usepackage{caption} 
\DeclareCaptionStyle{ruled}{labelfont=normalfont,labelsep=colon,strut=off} 
\usepackage{algorithm}
\usepackage{algorithmic}

\usepackage{amsmath,amssymb,amsfonts}
\usepackage{color}

\usepackage{subfigure}
\usepackage{amsfonts}
\usepackage{multirow}
\usepackage{multicol}
\usepackage{comment}
\newtheorem{definition}{Definition}
\newtheorem{theorem}{Theorem}
\newtheorem{lemma}{Lemma}
\newcommand{\methodname}{{\tt{mFairFL}}}

\usepackage{newfloat}
\usepackage{listings}
\floatstyle{ruled}
\newfloat{listing}{tb}{lst}{}
\floatname{listing}{Listing}
\title{Multi-dimensional Fair Federated Learning}
\usepackage{bibentry}

\author{
   Cong Su\textsuperscript{\rm 1},
    Guoxian Yu\textsuperscript{\rm 1,2},
    Jun Wang\textsuperscript{\rm 2},
    Hui Li\textsuperscript{\rm 1,2},
    Qingzhong Li\textsuperscript{\rm 1,2},
    Han Yu\textsuperscript{\rm 3}
}
\affiliations{
    \textsuperscript{\rm 1}School of Software, Shandong University, Jinan, China\\
    \textsuperscript{\rm 2}SDU-NTU Joint Centre for AI Research, Shandong University, Jinan, China\\
    \textsuperscript{\rm 3}School of Computer Science and Engineering, Nanyang Technological University, Singapore\\
    csu@mail.sdu.edu.cn, \{gxyu, kingjun, lih, lqz\}@sdu.edu.cn, han.yu@ntu.edu.sg
}

\begin{document}

\maketitle

\begin{abstract}
{
Federated learning (FL) has emerged as a promising collaborative and secure paradigm for training a model from decentralized data without compromising privacy. \emph{Group fairness} and \emph{client fairness} are two dimensions of fairness that are important for FL. Standard FL can result in disproportionate disadvantages for certain clients, and it still faces the challenge of treating different groups equitably in a population. The problem of privately training fair FL models without compromising the generalization capability of disadvantaged clients remains open. In this paper, we propose a method, called \methodname{}, to address this problem and achieve group fairness and client fairness simultaneously. \methodname{} leverages differential multipliers to construct an optimization objective for empirical risk minimization with fairness constraints. Before aggregating locally trained models, it first detects conflicts among their gradients, and then iteratively curates the direction and magnitude of gradients to mitigate these conflicts. Theoretical analysis proves \methodname{} facilitates the fairness in model development. The experimental evaluations based on three benchmark datasets show significant advantages of \methodname{} compared to seven state-of-the-art baselines. 
}
\end{abstract}

\section{Introduction}
The widespread adoption of machine learning models has given rise to significant apprehensions regarding fairness, spurring the emergence of fairness criteria and models. In recent times, a multitude of fairness criteria have been put forth, with one of the most widely acknowledged ones being \textbf{group fairness} \cite{hardt2016equality, ustun2019fairness}. Group fairness might also be mandated by legal statutes \cite{eu2012charter}, necessitating models to impartially treat distinct groups concerning sensitive attributes such as age, gender, and race.
Building upon these concepts of group fairness, numerous methodologies have been introduced to train equitable models, predicated on the premise that the model can directly access the complete training dataset \cite{zafar2017fairness, roh2021fairbatch}. However, the ownership of these datasets often resides with disparate institutions, rendering them inaccessible for sharing due to privacy safeguarding considerations. 

Federated learning (FL) \cite{wang2021field} stands as a distributed learning paradigm that facilitates the collective training of a model by multiple data custodians, all while retaining their data within their local domains. If each data steward was to individually train a fairness model on their own data and subsequently contribute it for aggregation, akin to the methods of FedAvg \cite{mcmahan2017communication} and FedOPT \cite{reddi2020adaptive}, a promising avenue emerges for augmenting model fairness within decentralized contexts. However, the presence of data heterogeneity, manifesting in variations in sizes and distributions across different clients, introduces a distortion to the localized efforts aimed at enhancing fairness in the global model. 

Consequently, a disparity emerges between the impartial model aggregated in a straightforward manner, utilizing fairness models trained on diverse client datasets, and the model achieved under centralized circumstances.
Meanwhile, a simplistic pursuit of minimizing the aggregation loss in the federated system can lead the global model astray, favoring certain clients excessively and disadvantaging others, thereby engendering what is termed as \emph{client fairness}. Preceding endeavors have predominantly centered on rectifying issues concerning client fairness. These efforts encompass methodologies such as re-weighting client aggregation weights \cite{zhao2022dynamic}, tackling distributed mini-max optimization challenges \cite{mohri2019agnostic}, or mitigating conflicts between clients \cite{hu2022federated}. 

In contrast, our emphasis pivots toward multi-dimensional fairness, encompassing both group and client fairness, aligning with legal stipulations and ethical considerations. This dual focus also significantly influences the willingness of clients to actively engage in the FL process, thereby contributing to datasets that are more comprehensive and representative for the training of the global model.
However, the inherent decentralized nature of this approach engenders complexities in achieving equitable training for a global model, particularly when confronted with the intricate tapestry of heterogeneous data distributions spanning the client landscape. The intricate challenge of privately training an equitable model from such decentralized, disparate data, while ensuring equitable treatment for each contributing client, poses a formidable conundrum. We aim to address this open and intricate quandary.

\begin{figure}[th!bp]
    \centering
    \includegraphics[width=7cm]{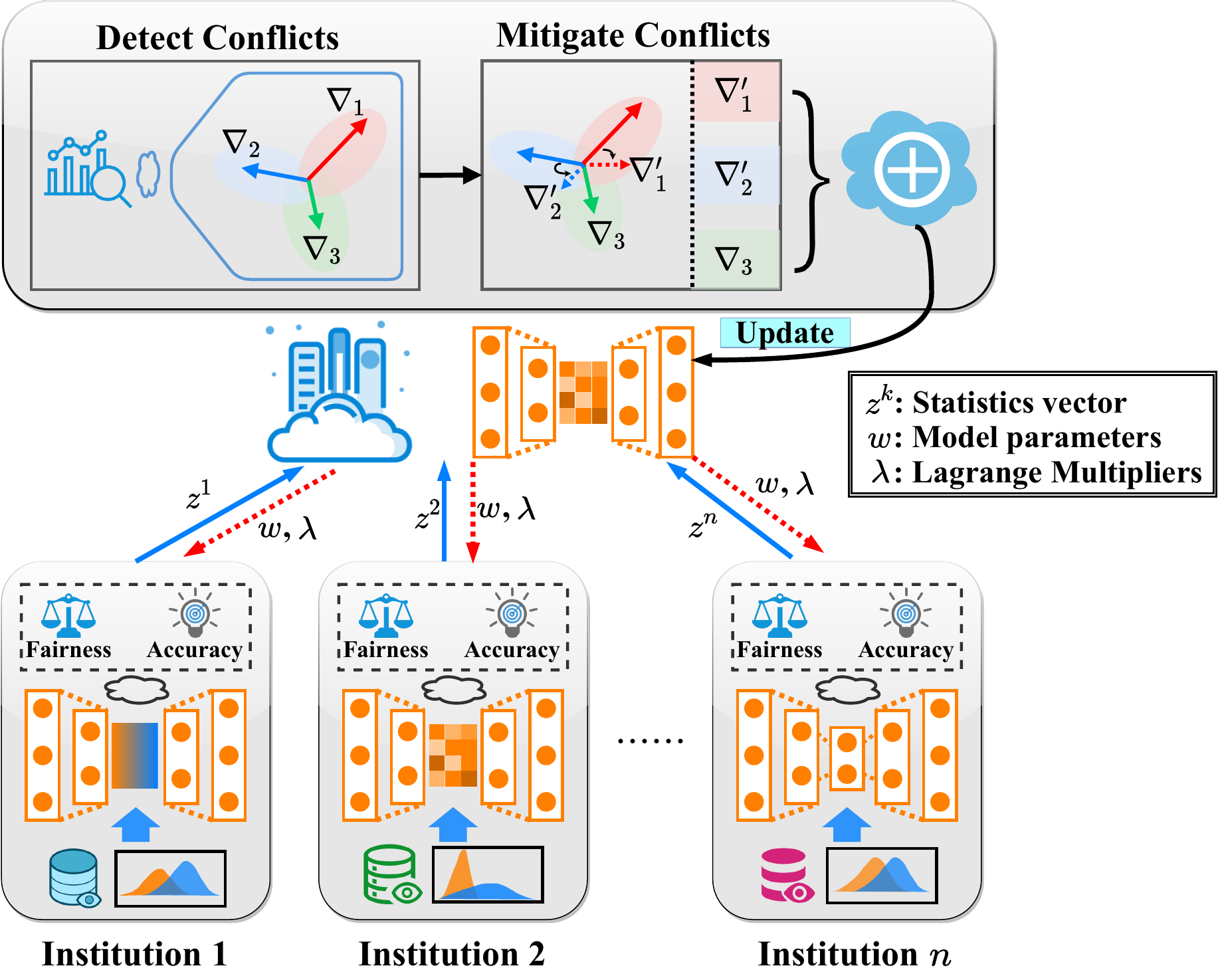}
    \caption{An overview of \methodname{}, which formulates a minimax constrained optimization problem in terms of accuracy and fairness. Before aggregating client gradients, it detects the presence of gradient conflicts, and then mitigates the conflicting gradients through gradient adjustments to align them with the overall fairness objective.}
    \vspace{-1.5em}
    \label{framework}
\end{figure}

We propose the \underline{M}ulti-dimensional \underline{Fair} \underline{F}ederated \underline{L}earning (\methodname{}) method, which aims to ensure equity not only among distinct sensitive groups but also across individual clients. The core principle of \methodname{} involves the optimization of client models under the guidance of fairness constraints. Prior to the execution of gradient aggregation on the central server, \methodname{} evaluates the potential presence of conflicting gradients among clients by assessing their gradient similarities.
Subsequently, \methodname{} undertakes an iterative process wherein it tactically adjusts the direction and magnitude of conflicting gradients to mitigate such disparities. Through this nuanced strategy, \methodname{} adeptly navigates the delicate balance between equitable treatment and optimal accuracy, catering to both marginalized sensitive groups and individual clients. The schematic framework of \methodname{} is depicted in Figure \ref{framework}.
Our contributions can be succinctly outlined as follows:\\
\noindent   (i) We introduce an innovative framework for fair federated learning, denoted as \methodname{}, and establish its capacity to bolster model fairness concerning sensitive groups within a decentralized data context.\\
\noindent    (ii) {\methodname{} conceptualizes the pursuit of fairness optimization through a meticulously designed minimax framework, replete with a group fairness metric as constraints. It analyzes and adjusts the trajectory and magnitude of potentially conflicting gradients throughout the training process, which adeptly augments group fairness across the entire populace while ensuring an impartial treatment of each client within the global model.}\\
\noindent   (iii) Through both theoretical and experimental analysis, we demonstrate that \methodname{} excels in mitigating gradient conflicts among clients, ultimately achieving a higher degree of group fairness compared to the state of the art (SOTA).


\section{Related Work}
With the growing concern surrounding fairness, various approaches have been proposed. To analyze the problem, we categorize fairness models into two types: centralized and federated, based on their training protocols.

\textbf{Fairness models on centralized data.}  
In the context of centralized data, it is common to modify the training framework to attain an appropriate level of group fairness, ensuring that a classifier exhibits comparable performance across different sensitive groups. Several techniques have been devised to address group fairness issues within the centralized setting, which can be categorized into three types: pre-processing, in-processing, and post-processing methods.
Pre-processing methods act before model training, aiming to eliminate implicit discrimination from datasets. For example, \citet{feldman2015certifying} made adjustments to sensitive attribute values in the dataset to achieve similar distributions among different sensitive groups. 
In-processing methods integrate fairness constraints directly into the model training process. \citet{garg2019counterfactual} added penalties to the loss function to minimize differences between real samples and their counterfactual counterparts. 
Post-processing methods focus on fairness through adjustments to the model's output. \citet{mishler2021fairness} employed double-robust estimators to reconstruct a trained model, aiming for approximate counterfactual fairness. Among these methods, in-processing techniques are often more effective in achieving a balance between model accuracy and fairness. 
The proposed \methodname{} is an in-processing fairness solution tailored for FL. 
For more extensive insights into fairness methods applied to centralized data, refer to the recent literature survey \cite{pessach2022review}.

\textbf{Fairness FL models on decentralized data.}  
In contrast, achieving fairness within the practical FL setting has received limited attention compared to centralized solutions \cite{wang2021federated}. The notion of `fairness' in FL differs slightly from the standard concept in centralized learning. \emph{Client fairness}, a popular fairness definition in FL, aims to ensure that all clients (i.e., data owners) achieve similar accuracy. 
Previous attempts to achieve client fairness in FL include modifying the aggregation weights of clients to achieve a uniform service quality for all clients \cite{li2019fair,zhao2022dynamic}, and combining minimax optimization and gradient normalization \cite{hu2022federated}. \citet{yue2022gifair} penalized the difference in aggregated loss to enforce consistent performance across participating clients. These fair FL solutions exclusively target client fairness. {Only a few studies are dedicated to group fairness in FL \cite{abay2020mitigating,du2021fairness}. For example, {\citet{galvez2021enforcing} distributively optimized local objective with fairness metric and then aggregated them to address fairness. \citet{zeng2021improving} updated the weight of local loss for each sensitive group during the global aggregation phase. \citet{ezzeldin2023fairfed} adapted client weights based on local fairness of each client and deviations from global one. However, these methods disregard the gradient conflicts, which lead to performance decline and unfavourable outcomes for certain clients.}

\methodname{} aims to eliminate bias towards different groups (group fairness) based on sensitive attributes and to learn a global model that benefits all clients, thereby achieving better client fairness alongside group fairness.


\section{Preliminaries}
\subsection{Federated Learning}
Following the typical FL setting \cite{mcmahan2017communication}, suppose there are $K$ different clients, and each client can only access its own dataset $\mathcal{D}_k=\{d_k^i=(s_k^i,x_k^i,y_k^i)\}_{i=1}^{n_k} \in \mathcal{D}$, where $s_k^i$ is the sensitive attribute of client $k$, $y_k^i$ is the label, $x_k^i$ is other observational attributes, $n_k$ is the number of client samples.
The goal of FL is to train a global model parameterized with $w \in \mathbb{R}^m$ ($m$ is the number of parameters) on client datasets $\mathcal{D}_k$ with guaranteed privacy. Formally, FL aims to solve:
\begin{equation}
    \min_{w \in \mathbb{R}^m} {\sum}_{i=1}^{K} p_i L(\mathcal{D}_i, w)
    \label{eq1}
\end{equation}
where $L(\mathcal{D}_i, w)=\frac{1}{n_i}\sum_{i=1}^{n_i}l(d^i_k,w)$ is the local objective function of client $i$ with weights $p_i \geq 0$, $\sum_{i=1}^K p_i=1$. 

\subsection{Fairness Notions}
{
Our goal is to train a global model in decentralized settings that satisfies group fairness with respect to sensitive attributes across all FL clients. We focus on three canonical group fairness notions, i.e., \emph{Demographic Parity}, \emph{Equalized Odds}, and \emph{Accuracy Parity} \cite{pessach2022review}. For the sake of exposition, we describe these notions in the centralized setting.

\begin{definition}[Demographic Parity]
The model's predictions $\hat{Y}$=$\hat{y}$ are statistically independent of the sensitive attribute $S$. The extent of a model's unfairness with respect to \emph{Demographic Parity} can be measured as follows:
\begin{equation}
    DP(\hat{y})=|\mathbb{P}[\hat{Y}=\hat{y}|S=s]-\mathbb{P}[\hat{Y}=\hat{y}]| \ \forall s \in S
    \label{eq3}
\end{equation}
\label{def1}
\vspace{-1em}
\end{definition}

\begin{definition}[Equalized Odds]
Given the label $Y$=$y$, The predictions $\hat{Y}$=$\hat{y}$ are statistically independent of the sensitive attribute $S$. i.e., for all $s \in S$ and $y \in Y$, we can measure the absolute difference between two prediction rates to quantify how unfair a model is in term of \emph{Equalized Odds}:
\begin{equation}
    EO(\hat{y}) = |\mathbb{P}[\hat{Y}=\hat{y}|S=s, Y=y]-\mathbb{P}[\hat{Y}=\hat{y}|Y=y]|
    \label{eq4}
\end{equation}
\label{def2}
\vspace{-1em}
\end{definition}

\begin{definition}[Accuracy Parity]
The model's mis-prediction rate is conditionally independent of the sensitive attribute. That is,
\begin{equation}
    \mathbb{P}[\hat{y} \neq y|S=s] = \mathbb{P}[\hat{y} \neq y] \ \forall s \in S
    \label{eq5}
\end{equation}
where, equivalently, we can measure the degree of unfairness in the model with respect to \emph{Accuracy Parity} as follows:
\begin{equation}
    AP(\hat{y})=|\mathbb{P}[L(\mathcal{D}, w)|S=s]-\mathbb{P}[L(\mathcal{D}, w)]|
    \label{eq6}
\end{equation}
where $L(\mathcal{D}, w)$ is the loss function minimized in problem \eqref{eq1}.
\label{def3}
\end{definition}

The above discussed fairness notions can be interpreted as the difference between each group and the overall population \cite{fioretto2020predicting}. Formally, these notions can be rewritten as:
\begin{equation}
    FN = |F(\mathcal{D},w)-F(\mathcal{D}^s,w)|
    \label{eq7}
\end{equation}
where $F(\mathcal{D},w) = \mathbb{E}_{d \sim \mathcal{D}}[f(d, w)]=\frac{1}{n}\sum_{d \in \mathcal{D}}f(d,w)$, $F(\mathcal{D}^s,w)=\frac{1}{n_s}\sum_{d^s \in \mathcal{D}}f(d^s,w)$. $\mathcal{D}^s$ is the subset of $\mathcal{D}$ with $S$=$s$, and $f$ is one of the fairness notions described above.
}


\subsection{Necessity for FL to improve fairness}
{
In this subsection, we analyse the advantage of FL for improving fairness in decentralized settings. 
To build a fair model in decentralized settings, an intuitive solution (hereon referred to as ``IndFair'') is to independently train the fair local model using client data. Specifically, for client $k$, IndFair trains a fair model by solving the following problem:
\begin{equation}
\begin{aligned}
    \min \ & L(\mathcal{D}_k, w) \\
    s.t. \ & |F(\mathcal{D}_k,w)-F(\mathcal{D}_k^s,w)| \leq \alpha, 
     \forall s \in S
\end{aligned}
    \label{eq8}
\end{equation}
where $\alpha$$\in$$[0,1]$ is the fairness tolerance threshold. Let $g_k^{\alpha}$ be the trained model of client $k$, then the overall performance of IndFair is defined as the mixture of all clients:
\begin{equation}
\begin{aligned}
    \mathrm{\hat{y}}|\mathrm{x},\mathrm{s} & \sim 
    \begin{cases}
        \mathrm{Bern}(g_1^{\alpha}(\mathrm{x,s})), & \mathrm{w.p.} \ 1/K \\
        \mathrm{Bern}(g_2^{\alpha}(\mathrm{x,s})), & \mathrm{w.p.} \ 1/K \\
        \cdots \cdots \\
        \mathrm{Bern}(g_{K}^{\alpha}(\mathrm{x,s})), & \mathrm{w.p.} \ 1/K
    \end{cases} \\
    & = \mathrm{Bern}((g_1^{\alpha}(\mathrm{x,s})+ \cdots +g_{K}^{\alpha}(\mathrm{x,s}))/K) \\
    & = \mathrm{Bern}(g_{\alpha}^{Seq})
\end{aligned}
\label{eq9}
\end{equation}
{where $\mathrm{Bern}$ stands for Bernoulli distribution, and $\mathrm{w.p.}$ is the abbreviation for `with probability'.}

On the other hand, we can train a fair global model (hereon referred to as ``FedFair'') on decentralized data through FL. The fair global model $g_{\alpha}^{Fed}$ is obtained by solving a constrained problem:
\begin{equation} 
\begin{aligned}
    \min & \ L(\mathcal{D}, w), \\ 
    s.t. \ & |F(\mathcal{D}_k,w)-F(\mathcal{D}_k^s,w)| \leq \alpha,  \\
    & \textit{for all} \quad k=1, 2,..., K.
\end{aligned}
\label{eq10}
\end{equation}
Here, an important question raises: \emph{can FedFair achieve a better fairness than IndFair?} The following theorem gives the confirm answer.
\begin{theorem}[Necessity for FL]
If the data distribution is highly heterogeneous across  clients, then $\min FN(g_{\alpha}^{Ind}) > \min FN(g_{\alpha}^{Fed})$.
\label{theo1}
\end{theorem}

Theorem \ref{theo1} means that in decentralized setting, there is a fairness gap between federated methods and non-federated ones, and FL improves the fairness performance. The proof is deferred into the Supplementary file. 
{
Nonetheless, when the centralized data are heterogeneous in terms of size and distribution across clients, adopting fairness-enhancing techniques in standard FL remains challenging, as it limits fairness improvement at the global level. In the next section, we introduce how \methodname{} alleviates client conflicts and trains a fair global model to bridge this gap.
}
}

\section{The Proposed \methodname{} Approach}
{
Theorem 1 demonstrates the potency of FL in effectively bolstering model fairness while safeguarding against data leakage within a decentralized context. Nevertheless, employing fairness methods directly within the FL framework might not be the optimal approach. This challenge arises from the significant heterogeneity in data distributions across clients. Consequently, the localized fairness performance could diverge from fairness across the entire population. Additionally, in this scenario, the concept of client fairness gains prominence as another critical facet of fairness that necessitates consideration.

To tackle these intricacies, we introduce \methodname{}, a solution designed to confidentially train a global model while integrating group fairness. This approach effectively mitigates the adverse effects of gradient conflicts among clients, as depicted in Figure \ref{framework}.
\methodname{} strategically transforms the fairness-constrained problem into an unconstrained problem that enforces fairness through the use of Lagrange multipliers. In each communication round, every client computes its training loss, measures of fairness violations, and gradients. Subsequently, these statistics are communicated to the FL server (aggregation phase). The server then identifies and rectifies conflicting gradients' direction and magnitude before aggregation. This refined model is then updated and distributed to clients (local training phase). This intricate process enables \methodname{} to attain a precise global model that remains equitable for both sensitive groups and individual clients. The subsequent subsections delve into the finer technical intricacies of our approach.

\subsection{The Local Training Phase}
Our goal is to train an optimal model from decentralized data while satisfying group fairness. For this purpose, we directly inject the group fairness constraint into the model training:
\begin{equation} 
\begin{aligned}
    & w^* = \min_{w \in \mathbb{R}^m} L(\mathcal{D}, w) \\
    & s.t. \ |F(\mathcal{D}, w)-F(\mathcal{D}^s, w)| \leq \alpha, \forall s \in S
\end{aligned}
\label{eq11}
\end{equation}
{where $L(\mathcal{D}, w)=\frac{1}{n}\sum_{d \in \mathcal{D}}l(d,w)$, $F(\mathcal{D}, w)$ is the fairness metric defined in Eq. \eqref{eq7}.}

Let $\mathbf{h}(w)$=$[h_1(w),h_2(w),\cdots,h_{|S|}(w)]$ where $h_s(w)$=$|F(\mathcal{D}, w)-F(\mathcal{D}^s, w)|-\alpha$. We use the similar technique from the Lagrangian approach \cite{fioretto2021lagrangian} to relax the constraint:
\begin{equation}
    J(w,\mathbf{\alpha})=L(\mathcal{D}, w) + \mathbf{\lambda} \mathbf{h}(w).
    \label{eq12}
\end{equation}
The relaxation provides more freedom for the optimization algorithm to find solutions that may not strictly satisfy all the constraints, but rather approximate them within an acceptable range.

Thus, the objective function in Eq. \eqref{eq12} can be optimized using gradient descent/ascent:
\begin{equation}
    \begin{cases}
        \mathbf{\lambda} \gets \mathbf{\lambda} + \gamma \mathbf{h}(w), \\
        w \gets w - \eta (\nabla_w L(\mathcal{D}, w) + \mathbf{\lambda} \nabla_w \mathbf{h}(w)).
    \end{cases}
\label{eq13}
\end{equation}

Based on Eq. \eqref{eq13}, each client computes the following statistics required for the server to perform model updates:
\begin{equation}
\begin{aligned}
    & L(\mathcal{D}, w); \nabla_w L(\mathcal{D}, w); F(\mathcal{D}, w); [F(\mathcal{D}^s, w)]_{s \in S}; \\
    & \nabla_w F(\mathcal{D}, w); \nabla_w [F(\mathcal{D}^s, w)]_{s \in S}.
\end{aligned}
    \label{eq14}
\end{equation}
In fact, some of these statistics can be obtained from others: $F(\mathcal{D}, w)$=$\sum_{s \in S}F(\mathcal{D}^s, w)$, $\nabla_w F(\mathcal{D}, w)$=$\sum_{s \in S}\nabla_w F(\mathcal{D}^s, w)$. Therefore, in each communication round, a client reports a statistics vector to the server as:
\begin{equation}
\begin{aligned}
    z_k = & [L(\mathcal{D}_k, w); \nabla_w L(\mathcal{D}_k, w); [F(\mathcal{D}_k^s, w)]_{s \in S}; \\
    & \nabla_w [F(\mathcal{D}_k^s, w)]_{s \in S}]
\end{aligned}
    \label{eq15}
\end{equation}

We define the training loss of client $k$ in round $t$ as $l_k^t$=$L(\mathcal{D}_k, w)+\mathbf{\lambda}\mathbf{h}(w)$, and the updated gradient  $g_k^t$=$\nabla_w L(\mathcal{D}_k, w)+\mathbf{\lambda}\nabla_w\mathbf{h}(w)$. Let $G_t$=$\{g_1^t, g_2^t,\cdots, g_K^t\}$ represent the gradients received by the server from clients, and $L_t$=$\{l_1^t, l_2^t, \cdots, l_K^t\}$ be the received client losses.

\subsection{The Aggregation Phase}
During the aggregation phase, the server leverages the information provided by clients to refine and update the global model. Owing to the presence of diverse data distributions, gradient conflicts emerge among clients. In isolation, these conflicts might not be inherently detrimental, as straightforward gradient averaging can effectively optimize the global objective function \cite{mcmahan2017communication}. However, when conflicts among gradients involve considerable variations in magnitudes, certain clients could encounter pronounced drops in performance.
For instance, consider the scenario of training a binary classifier. If a subset of clients holds a majority of data pertaining to one class, and conflicts in gradients arise between these two classes, the global model could become skewed toward the majority-class clients, thereby compromising performance on the other class. Moreover, even when class balance is maintained among clients, disparities in gradient magnitudes may persist due to divergent sample sizes across clients.

Therefore,  before aggregating clients' gradients in each communication round, \methodname{} first checks whether there are any conflicting gradients among clients. If there are gradient conflicts, then at least a pair of client gradients $(g_i^t, g_j^t)$ such that $\cos(g_i^t, g_j^t)<\hat{\phi}_{ij}^t$, where $\hat{\phi}_{ij}^t \geq 0$ is the gradient similarity goal of $t$-th communication round. Note that interactions among gradients (i.e., gradient similarity goal) change significantly across clients and communication rounds. Thus, \methodname{} performs Exponential Moving Average (EMA) to set appropriate gradient similarity goals for clients $i$ and $j$ in round $t$:
\begin{equation}
    \hat{\phi}_{ij}^t = \delta \hat{\phi}_{ij}^{t-1} + (1-\delta)\phi_{ij}^t
    \label{eq16}
\end{equation}
where  $\delta$ is the hyper-parameter, and $\phi_{ij}^t=\cos(g_i^t, g_j^t)$ is the computed gradient similarity. Specifically, $\hat{\phi}_{ij}^0 = 0$.

In order to mitigate the adverse repercussions stemming from gradient conflicts among clients, \methodname{} introduces an innovative gradient aggregation strategy. Specifically, the approach initiates by arranging clients' gradients within $G_t$ in ascending order, based on their respective loss values. This orchestrated arrangement yields $PO_t$, which outlines the sequence for utilizing each gradient as a reference projection target. Subsequently, through an iterative process, \methodname{} systematically adjusts the magnitude and orientation of the $k$-th client gradient, denoted as $g_k^t$, so as to align with the desired similarity criteria between $g_k^t$ and the target gradient $g_j^t \in PO_t$, in accordance with the prescribed order set by $PO_t$:
\begin{equation}
    g_k^{t}=c_1\cdot g_k^{t} + c_2 \cdot g_j^t
    \label{eq17}
\end{equation}
Since there are infinite valid combinations of $c_1$ and $c_2$, we fix $c_1=1$ and apply the Law of Sines on the planes of $g_k^t$ and $g_j^t$ to calculate the value of $c_2$, and obtain the derived new gradient for the $k$-th client:
\begin{equation}
    g_k^{t}=g_k^{t}-\frac{||g_k^{t}||(\phi_{kj}^t \sqrt{1-(\hat{\phi}_{kj}^t)^2}-\hat{\phi}_{kj}^t\sqrt{1-(\phi_{kj}^t)^2})}{||g_j^t||\sqrt{1-(\hat{\phi}_{kj}^t)^2}} \cdot g_j^t
    \label{eq18}
\end{equation}
The derivation detail is deferred into the Supplementary file.

\begin{theorem}
Suppose there is a set of gradients $G=\{g_1,g_2,...,g_K\}$ where $g_i$ always conflicts with $g_j^{t_j}$ before adjusting $g^{t_j}_j$ to match similarity goal between $g^{t_j}_j$ and $g_i$ ($g^{t_j}_j$ represents the gradient adjusting $g_j$ with the target gradients in $G$ for $t_j$ times). Suppose $\epsilon_1 \leq |cos(g^{t_i}_i, g^{t_j}_j)| \leq \epsilon_2$, $0$$<$$\epsilon_1$$<$$\hat{\phi}_{ij}$$\leq$$\epsilon_2$$\leq 1$, for each $g_i \in G$, as long as we iteratively project $g_i$ onto $g_k$'s normal plane (skipping $g^i$ itself) in the ascending order of $k$=$1,2,\cdots,K$, the larger the $k$ is, the smaller the upper bound of conflicts between the aggregation gradient of global model $g^{global}$ and $g_k$ is. The maximum value of $|g^{global} \cdot g_k|$ is bounded by $\frac{K-1}{K}(\max_i||g_i||)^2\frac{\epsilon_2X_{\max}(1-X_{\min})(1-(1-X_{min})^{K-k})}{X_{\min}}$, where $X_{\max}$=$\frac{\epsilon_2\sqrt{1-\hat{\phi}^2}-\hat{\phi}\sqrt{1-\epsilon_2^2}}{\sqrt{1-\hat{\phi}^2}}$ and $X_{\min}$=$\frac{\epsilon_1\sqrt{1-\hat{\phi}^2}-\hat{\phi}\sqrt{1-\epsilon_1^2}}{\sqrt{1-\hat{\phi}^2}}$.
\label{theo2}
\end{theorem}

Theorem \ref{theo2} substantiates that the later a client's gradient assumes the role of the projection target, the fewer conflicts it will engage in with the ultimate averaged gradient computed by \methodname{}. Consequently, in the pursuit of refining the model's performance across clients with comparatively lower training proficiency, we position clients with higher training losses towards the end of the projecting target order list, denoted as $PO_t$.
Additionally, these gradients provide the optimal model update direction. To further amplify the focus on \emph{client fairness}, we permit $\beta K$ clients with suboptimal performance to retain their original gradients. The parameter $\beta$ modulates the extent of conflict mitigation and offers a means to strike a balance. When $\beta$=1, all clients are mandated to mitigate conflicts with others. Conversely, when $\beta$=0, all clients preserve their original gradients, aligning \methodname{} with FedAvg. By adopting this approach, \methodname{} effectively alleviates gradient conflicts, corroborated by the findings in Theorem \ref{theo2}. Consequently, \methodname{} is equipped to set an upper limit on the maximum conflict between any client's gradient and the aggregated gradient of the global model. This strategic stance enables \methodname{} to systematically counteract the detrimental repercussions stemming from gradient conflicts. Algorithm 1 in the Supplementary file outlines the main procedures of \methodname{}.

Theorem \ref{theo3} proves that \methodname{} can find the optimal value $w^*$ within a finite number of communications. This explains why \methodname{} can effectively train a group and client fairness-aware model in the decentralized setting. The proof can be found into the Supplementary file.
\begin{theorem}
Suppose there are $K$ objective functions $J_1(w), J_2(w),\cdots, J_K(w)$, and each objective function is differentiable and L-smooth. Then \methodname{} will converage to the optimal $w^*$ within a finite number of steps.
\label{theo3}
\end{theorem}
}

\section{Experimental Evaluation}

\subsection{Experimental Setup}
\label{Exp_Setup}
In this section, we conduct experiments to evaluate the effectiveness of \methodname{} using three real-world datasets: Adult  \cite{lichman2017uci}, COMPAS  \cite{ProPublica2021compas}, and Bank \cite{moro2014data}. The Adult dataset contains 48,842 samples, with `gender' treated as the sensitive attribute. There are 7,214 samples in the COMPAS dataset, with `gender' treated as the sensitive attribute. As for the Bank dataset with 45,211 samples, with `age' treated as the sensitive attribute. We split the data among five FL clients in an non-iid manner.\footnote{Due to page limit, we include the experiments conducted in a more general setting with multiple sensitive attributes and multiple values for each sensitive attribute in Supplementary file.} 

For the purpose of comparative analysis, we consider several baseline methods, categorized into three groups: (i) independent training of the fair model within a decentralized context (IndFair); (ii) fair model training via FedAvg (FedAvg-f); (iii) fair model training within a centralized setting (CenFair). Three SOTA FL with group fairness: (i) FedFB \cite{zeng2021improving}, which adjusts each sensitive group's weight for aggregation; (ii) FPFL \cite{galvez2021enforcing}, which enforces fairness by solving the constrained optimization; (iii) FairFed \cite{ezzeldin2023fairfed}, which adjusts clients' weights based on locally and global trends of fairness mtrics. In addition to these, we evaluate our proposed \methodname{} against cutting-edge FL methods that emphasize \emph{client fairness}, including: (i) q-FFL \cite{li2019fair}, which adjusts client aggregation weights using a hyperparameter $q$; (ii) DRFL \cite{zhao2022dynamic}, which automatically adapts client weights during model aggregation; (iii) Ditto \cite{li2021ditto}, a hybrid approach that merges multitask learning with FL to develop personalized models for each client; and (iv) FedMGDA+ \cite{hu2022federated}, which frames FL as a multi-objective optimization problem.
Throughout our experiments, we adhere to a uniform protocol of 10 communication rounds and 20 local epochs for all FL algorithms. For other methods, we execute 200 epochs, leveraging cross-validation techniques on the training sets to determine optimal hyperparameters for the comparative methods. All algorithms are grounded in ReLU neural networks with four hidden layers, thereby ensuring an equal count of model parameters.\footnote{Further elaboration on the selection of hyperparameters for \methodname{} can be found in the Supplementary file.} We use the same server (Ubuntu 18.04.5, Intel Xeon Gold 6248R and Nvidia RTX 3090) to perform experiments.

\begin{table*}[!t]
    \centering
    \resizebox{0.9\linewidth}{!}{
    \renewcommand\arraystretch{1.1}
    \begin{tabular}{r|l l l l l|l l l l l|l l l l l}
        \hline
        \hline
        \multirow{2}{*}{} & \multicolumn{5}{c|}{Adult} & \multicolumn{5}{c|}{Compas} & \multicolumn{5}{c}{Bank} \\
        \cline{3-5}\cline{8-10}\cline{13-15}
        & Acc. & DP & EO & AP & CF & Acc. & DP & EO & AP & CF & Acc. & DP & EO & AP & CF \\
        \hline
        IndFair & .768$\bullet$ & .083$\bullet$ & .071$\bullet$ & .077$\bullet$ & - & .573$\bullet$ & .083$\bullet$ & .097$\bullet$ & .085$\bullet$ & - & .831 & .028$\bullet$ & .025$\bullet$ & .029$\bullet$ & - \\
        FedAvg-F & .706$\bullet$ & .224$\bullet$ & .164$\bullet$ & .218$\bullet$ & .232$\bullet$ & .558$\bullet$ & .059$\bullet$ & .066$\bullet$ & .062$\bullet$ & .184$\bullet$ & .828$\bullet$ & .033$\bullet$ & .034$\bullet$ & .033$\bullet$ & .143$\bullet$ \\
        \hline
        FedFB & .779 & .014 & .007 & .011 & .058$\bullet$ & .557$\bullet$ & .023$\bullet$ & .019$\bullet$ & .021$\bullet$ & .033 & .837 & .014$\bullet$ & .016$\bullet$ & .009 & .067$\bullet$ \\
        FPFL & .754$\bullet$ & .023$\bullet$ & .016$\bullet$ & .019$\bullet$ & .228$\bullet$ & .553$\bullet$ & .033$\bullet$ & .018$\bullet$ & .024$\bullet$ & .157$\bullet$ & .822$\bullet$ & .008 & .012$\bullet$ & .010$\bullet$ & .153$\bullet$ \\
        FairFed & .756$\bullet$ & \bf .009 & .004 & .008 & .244$\bullet$ & .551$\bullet$ & .009 & \bf .003 & .004 & .186$\bullet$ & .824$\bullet$ & .003 & .004 & .004 & .164$\bullet$ \\
        FedMGDA+ & \bf .837$\circ$ & .238$\bullet$ & .237$\bullet$ & .238$\bullet$ & .063$\bullet$ & \bf .635$\circ$ & .136$\bullet$ & .141$\bullet$ & .137$\bullet$ & .044$\bullet$ & \bf .874$\circ$ & .084$\bullet$ & .077$\bullet$ & .085$\bullet$ & .065$\bullet$ \\
        \hline
        CenFL & .812$\circ$ & .014 & .008 & .013 & - & .616 & .014$\bullet$ & .008 & .011$\bullet$ & - & .866$\circ$ & \bf .001 & \bf .000 & \bf .002 & - \\
        \hline
        \hline
        \textbf{mFairFL} & .792 & .012 & \bf .003 & \bf .007 & \bf .036 & .596 & \bf .005 & .009 & \bf .003 & \bf .022 & .844 & .005 & .006 & .003 & \bf .028 \\
        \hline
        
    \end{tabular}
    }
    \vspace{-1em}
    \caption{Accuracy and the violation of \emph{Group fairness} and \emph{Client Fairness} results on the three datasets with high data heterogeneity among clients. The best results in fairness are highlighted in \textbf{boldface}. $\circ/\bullet$  indicates that \methodname{} is statistically worse/better than the compared method by student pairwise $t$-test at 95\% confident level. `-' implies not applicable.}
    \label{table2}
    \vspace{-1em}
\end{table*}

\subsection{Estimation on Group fairness}
\label{EoDP}
{\color{blue}

}
We undertake a comprehensive comparative analysis, focusing on the accuracy and group fairness aspects of the evaluated methods. To delve into the intricate relationship between method performance and data heterogeneity, we adopt a random assignment strategy. Specifically, we randomly assign 30\%, 30\%, 20\%, 10\%, 10\% of the samples from group 0 and 10\%, 20\%, 20\%, 20\%, 30\% of the samples from group 1 to five clients, respectively. The outcomes of this data splitting strategy, encompassing average accuracy along with standard deviations and the Demographic Parity violation score for each method, are outlined in Table S2. Furthermore, for datasets characterized by pronounced data heterogeneity, we draw samples from each group across five clients at a ratio of 50\%, 10\%, 10\%, 20\%, 10\%, and 10\%, 40\%, 30\%, 10\%, 10\%, respectively. The corresponding experimental outcomes are showcased in Table \ref{table2}.
From the insights gleaned from Tables S2 and \ref{table2}, we observe that:\\
\noindent (i) \methodname{} prominently enhances fairness, achieving a parity of fairness akin to CenFair. This substantiates \methodname{}'s efficacy in skillfully training fair models for sensitive groups within the decentralized data landscape.\\
\noindent (ii) The lackluster performance of IndFair in terms of group fairness accentuates that in a decentralized scenario, fairness models exclusively trained on local data fall considerably short of achieving group fairness at a population-wide level. It is also noteworthy that FedAvg-f occasionally exhibits lower accuracy than IndFair. This discrepancy arises from the aggregation strategy of FedAvg-f, which can have unintended consequences for certain clients, causing the averaged local fair models to not be equitable for any sub-distribution. Conversely, IndFair manages to ensure fairness for specific sub-distributions through the training of each local model.

\noindent (iii) {In direct comparison, \methodname{} distinctly outperforms FedAvg-f in both accuracy and fairness, thus underscoring the constraints inherent in merely grafting fairness techniques onto the FL paradigm. Evidently, the group fairness achieved by FedAvg-f lags behind the fairness exhibited across the entire population. This gap is particularly pronounced in scenarios characterized by high data heterogeneity among clients. Through the judicious amalgamation of fairness techniques with the decentralized essence of FL, and its steadfast commitment to ensuring advantageous model updates for all clients, \methodname{} adeptly enhances both the overarching fairness and accuracy, thereby offering a comprehensive improvement.} 

\noindent (iv) Notably, \methodname{} can better trade-off accuracy and group fairness than FedFB, FPFL and FairFed. This is because they overlook the detrimental effects of the conflicting gradients with large difference in the magnitudes, leading to accuracy reduction and harming certain clients. FedMGDA+ frequently yields the highest accuracy in various scenarios, but markedly infringes upon the fairness of model decisions as applied to disadvantaged groups. This is primarily attributed to the fact that FedMGDA+ concentrate solely on aligning client accuracy without due regard for mitigating discrimination against sensitive groups.

\noindent (v) {Upon juxtaposing the outcomes presented in Table \ref{table2} (high heterogeneity) with those in Table S2 (low heterogeneity), a salient observation arises: \methodname{} demonstrates a marginal decrease in both fairness and accuracy when transitioning from low to high data heterogeneity. This underscores the robustness intrinsic to \methodname{} when grappling with heterogeneous data. Such a consistency aligns with our initial expectations, as \methodname{} adeptly orchestrates gradient directions and magnitudes to navigate conflicts, thereby ensuring equitable model updates across all clients. Conversely, FedAvg-f manifests notable performance disparities across distinct data heterogeneity levels, with a particularly steep decline observed in scenarios characterized by high data heterogeneity. This vulnerability is attributed to FedAvg-f's simplistic gradient averaging approach, which insufficiently accommodates the intricate impact of data heterogeneity on the global model.}
\begin{table*}[!t]
    \centering
    \resizebox{0.6\linewidth}{!}{
    \begin{tabular}{r r| c|| c c c c |c }
        \hline
        \multicolumn{2}{c|}{} &FedAvg &q-FFL &DRFL &Ditto &FedMGDA+ &\textbf{\methodname{}} \\
        \hline
        \multirow{2}*{Adult}
        & Accuracy & .792$\bullet$ & .718$\bullet$ & .762$\bullet$ & .834$\bullet$ & \bf .837 & .853 \\
        ~ & CF Vio. & .219$\bullet$ & .081$\bullet$ & .084$\bullet$ & .074$\bullet$ & .063$\bullet$ & \bf .035  \\
        \hline
        \multirow{2}*{Compas}
        & Accuracy & .594$\bullet$ & .566$\bullet$ & .598$\bullet$ & .629$\bullet$ & .635$\bullet$ & \bf .668 \\
        ~ & CF Vio. & .179$\bullet$ & .058$\bullet$ & .031$\bullet$ & .024 & .044$\bullet$ & \bf .018  \\
        \hline
        \multirow{2}*{Bank}
        & Accuracy & .842$\bullet$  & .816$\bullet$ & .864$\bullet$ & .877 & .874$\bullet$ & \bf .889 \\
        ~ & CF Vio. & .147$\bullet$ & .110$\bullet$ & .090$\bullet$ & .068$\bullet$ & .065$\bullet$ & \bf .022  \\
        \hline   
    \end{tabular}
    }
    \vspace{-1em}
    \caption{Accuracy ($\uparrow$) and \emph{Client Fairness} violation score ($\downarrow$) on three datasets with high heterogeneity among clients. $\circ/\bullet$  indicates that \methodname{} is statistical worse/better than the compared method (student pairwise $t$-test at 95\% confident level).}
    \label{table3}
    \vspace{-1em}
\end{table*}

\subsection{Estimation on Client Fairness}
\label{EoCF}
The group fairness-aware model cultivated by \methodname{} brings about advantages for each participating client, all while avoiding any undue preference towards specific clients. To further validate this assertion, we embark on a series of experiments designed to evaluate \methodname{}'s performance in terms of \emph{Client Fairness}, subsequently juxtaposing it against other pertinent fairness methods. \emph{Client Fairness} stands as a potent metric for gauging whether the global model disproportionately favors particular clients while disregarding the rest. To further accentuate the discerning capabilities of \methodname{}, we undertake the random allocation of samples: 50\% and 10\% of group 0 samples, coupled with 10\%, 20\%, and 10\% of group 1 samples, are assigned to the 1st, 2nd, 3rd, 4th, and 5th clients, respectively. This deliberate strategy accentuates pronounced data heterogeneity across clients. For reference, FedAvg constitutes the baseline in this experimental setup. The resulting accuracy and violation scores pertinent to \emph{Client Fairness} for each method are succinctly presented in Table \ref{table3}. Our observations from this comparative analysis are as follows:\\
\noindent (i) Remarkably, \methodname{} emerges as the frontrunner, boasting the most modest client fairness violation scores while achieving accuracy on par with other fairness-focused FL methods. In essence, \methodname{} excels in abating the potential biases inherent to FL contexts. Through its meticulous consideration of conflicting gradients and adept adjustments to their directions and magnitudes, \methodname{} guarantees a more equitable distribution of model updates among clients. This concerted effort tangibly diminishes the breach of client fairness, ultimately heralding a more even allocation of model updates among all participating clients.
In contrast, both q-FFL and DRFL endeavor to tackle client fairness by manipulating client aggregation weights, but falter in effectively addressing conflicts characterized by substantial gradient magnitude disparities. Ditto aims to strike a balance between local and global models, engendering personalized models for individual clients. However, its global model aggregation strategy closely resembles that of FedAvg, potentially yielding unfavorable outcomes for certain clients. In the same vein, FedMDGA+ aspires to pinpoint a shared update direction for all clients during federated training, inadvertently overlooking the influential role played by gradient magnitudes in model aggregation. Therefore, it is evident that \methodname{} stands as the epitome of achievement, outperforming its counterparts both in terms of client fairness and accuracy.\\
\noindent (ii) FedAvg, unfortunately, languishes at the bottom of the performance spectrum, marked by inferior accuracy and client fairness. This regression is traceable to its rudimentary averaging strategy, which disregards the disparate contributions of individual clients. Consequently, when confronting gradients in conflict with significantly divergent magnitudes, FedAvg becomes susceptible to overfitting certain clients at the detriment of others. The significant difference in performance between \methodname{} and FedAvg demonstrates the potency of \methodname{} in counteracting client conflicts.\\
\noindent (iii) To further solidify \methodname{}'s efficacy, we furnish the number of iterations imperative for all compared methods to attain their optimal performance in Figure S1 of the Supplementary file. This visual depiction affords insights into the convergence trajectories undertaken by distinct methods over iterations. Notably, \methodname{} exhibits commendable performance levels and converges towards the pinnacle of client fairness within a noticeably fewer (or comparable) count of communication rounds. This clearly underscores \methodname{}'s capacity to efficiently train the model, attaining the desired accuracy and client fairness benchmarks with commendable efficacy.

\subsection{Ablation Study}
\label{ablation}
To prove the necessity of the projection order of \methodname{}, we introduce two variants of \methodname{}: (i) \methodname{}-rnd adjusts the gradients in a random order of the projection target. (ii) \methodname{}-rev adjusts gradients in the opposite order. The experimental settings are the same as the previous subsection. The results are shown in Table \ref{table4}.
In can be observed that the projection order has significant impact on the effectiveness of gradient projection. \methodname{}-rnd ignores the information provided by client losses. Thus, it loses to \methodname{} in terms of \emph{group fairness} and \emph{client fairness}. \methodname{}-rev achieves lower fairness than \methodname{}-rnd, indicating that the global model tends to neglecting clients with poorer performance when adjusting gradients in the opposite order of \methodname{}. The best multi-dimensional fairness performance is obtained by \methodname{}. This confirms that its loss-based order helps improve fairness.

\begin{table}[h!tbp]
    \centering
    \resizebox{0.9\linewidth}{!}{
    \begin{tabular}{r r| l l ||l }
        \hline
        \multicolumn{2}{c|}{} &\methodname{}-rnd &\methodname{}-rev &\textbf{\methodname{}} \\
        \hline
        \multirow{4}*{Adult}
        & Accuracy & .774  & .768$\bullet$ & \bf .792  \\
        ~ & DP Vio. & .018 & .027$\bullet$ & \bf .012  \\
        ~ & EO Vio. & .013$\bullet$ & .026$\bullet$ & \bf .003  \\
        ~ & AP Vio. & .017$\bullet$ & .028$\bullet$ & \bf .007  \\
        ~ & CF Vio. & .049$\bullet$ & .064$\bullet$ & \bf .036  \\
        \hline
        \multirow{4}*{Compas}
        & Accuracy & .577$\bullet$  & .580 & \bf .596  \\
        ~ & DP Vio. & .014$\bullet$ & .020$\bullet$ & \bf .005  \\
        ~ & EO Vio. & .015$\bullet$ & .018$\bullet$ & \bf .009  \\
        ~ & AP Vio. & .012$\bullet$ & .019$\bullet$ & \bf .003  \\
        ~ & CF Vio. & .044$\bullet$ & .049$\bullet$ & \bf .022  \\
        \hline
        \multirow{4}*{Bank}
        & Accuracy & .837  & .822$\bullet$ & \bf .844  \\
        ~ & DP Vio. & .009 & .023$\bullet$ & \bf .005  \\
        ~ & EO Vio. & .013 & .019$\bullet$ & \bf .006  \\
        ~ & AP Vio. & .007 & .021$\bullet$ & \bf .003  \\
        ~ & CF Vio. & .047$\bullet$ & .077$\bullet$ & \bf .028  \\
        \hline
        
    \end{tabular}
    }
    \vspace{-1em}
    \caption{Accuracy ($\uparrow$), \emph{Group fairness} and \emph{Client Fairness} violation scores ($\downarrow$) of \methodname{} and its variants. $\bullet$ indicates that \methodname{} is statistical better than the variant (student pairwise $t$-test at 95\% confident level).}
    \label{table4}
    \vspace{-1em}
\end{table}

\section{Conclusions}
{
Addressing both group fairness and client fairness is paramount when considering fairness issues in the realm of FL. This paper introduces the novel \textbf{\methodname{}} method as a groundbreaking solution that adeptly navigates these dual dimensions of fairness. \methodname{} formulates the optimization conundrum as a minimax problem featuring group fairness constraints. Through meticulous adjustments to conflicting gradients throughout the training regimen, \methodname{} orchestrates model updates that distinctly benefit all clients in an equitable manner. 
Both theoretical study and empirical results confirm that mitigating client conflicts during global model update improves the fairness for sensitive groups, and \methodname{} effectively achieves both group fairness and client fairness.
}

\clearpage

\appendix   
\setcounter{table}{0}   
\setcounter{figure}{0}
\setcounter{section}{0}
\setcounter{equation}{0}
\setcounter{theorem}{0}

\renewcommand{\thetable}{S\arabic{table}}
\renewcommand{\thefigure}{S\arabic{figure}}
\renewcommand{\thesection}{S\arabic{section}}
\renewcommand{\theequation}{S\arabic{equation}}

\twocolumn[
\begin{@twocolumnfalse}
	\section*{\centering{Multi-dimensional Fair Federated Learning \\
Supplementary file \\[25pt]}}
\end{@twocolumnfalse}
]

\section{Algorithm table}
The comprehensive procedure of \methodname{} is delineated in Algorithm \ref{alg1}. Lines 1-2 initialize model parameters $w_0$, $\mathbf{\lambda}_0$, and EMA variables. Lines 3-7 calculate the statistic vector for each client and upload them to the server. Line 8 updates Lagrange multiplier. Lines 9-11 curate the conflicting gradients, and then aggregates them. Lines 12-13 update global model and broadcasts model parameters $w$, $\mathbf{\lambda}$ to clients. Algorithm \ref{alg2} presents the details of how the gradient conflict mitigation is performed in Line 10 of Algorithm \ref{alg1}. Lines 1-2 select the clients requiring gradient adjustment and initialize the adjusted gradients. Lines 3-8 detect gradient conflicts and correct the magnitude and direction of conflicting gradients. Lines 9-12 update EMA variables and aggregate gradients.

\begin{algorithm}[!b]
\caption{\methodname{}}
\label{alg1}
\textbf{Input}: The number of clients $K$, the set of clients' datasets $\{\mathcal{D}^k\}_{i=1}^K$, learning rates for $w$ and $\mathbf{\lambda}$, $\gamma$ and $\eta$, respectively, the maximum communication round $T$, fairness toleration threshold $\alpha$, gradient projection rate $\beta$, and EMA decay $\delta$. \\
 \textbf{Output}: Fairness model $w^*$
\begin{algorithmic}[1] 
\STATE Randomly initialize the global model $w_0$ and set $\mathbf{\lambda}_0=0$ on the server side.
\STATE Initialize EMA variables $\hat{\phi}_{ij}^0=0$, $\forall i, j \in \{1,\cdots,K\}$
\FOR{$t=1,2,\cdots,T$}
\STATE 
$\rhd$ Procedure at clients\\
\FOR{each client $k$ in parallel}
\STATE Client $k$ calculates and uploads a statistics vector $z_k$ via. Eq. (14)
\ENDFOR
\\
$\rhd$ Procedure at the server\\
\STATE 
$\mathbf{\lambda}_t=\mathbf{\lambda}_{t-1}+\gamma \mathbf{h}(w)$
\STATE Sort the clients' gradients $G_t$ into a projecting order list $PO_t=[g_1^t,g_2^t,...,g_K^t]$ with clients' loss $L_t$ in ascending order.
\STATE $g_t^{global}, \hat{\phi}^t \gets \mathrm{DiminishConflicts}(PO_t, G_t, \beta, \delta, \hat{\phi}^{t-1})$
\STATE $g_t^{global} \gets \frac{g_t^{global}}{||g_t^{global}||} \cdot ||\frac{1}{m}\sum_{i=1}^K g_i^t||$
\STATE Update the global model $w_t$ $\gets$ $w_{t-1}$-$\eta$ $g_t^{global}$
\STATE Broadcast $w_{t}$ and $\mathbf{\lambda}_{t}$ to clients\\
\ENDFOR
\end{algorithmic}
\end{algorithm}

\begin{algorithm}[!t]
\caption{DiminishConflicts}
\label{alg2}
\textbf{Input}: The projecting order list $PO_t$, the clients' gradients list $G_t$, gradient projecting rate $\beta$, EMA decay $\delta$, EMA variables $\hat{\phi}^{t-1}$ \\
\textbf{Output}: $g_t^{global}$, $\hat{\phi}^t$
\begin{algorithmic}[1] 
\STATE Select the set $P_t^{\beta}$ of the top $\beta$ clients in $PO_t$
\STATE Set $g_k^{PC} \gets g_k^t$ for each client $k=1,2,...,K$
\FOR{each client $k \in P_t^{\beta}$}
\FOR{each $g_i^t \in PO_t$}
\STATE $\phi_{ki}^{t-1} \gets \frac{g_k^{PC} \cdot g_i^t}{||g_k^{PC}|| ||g_i^t||}$
\IF{$\phi_{ki}^{t-1} < \hat{\phi}_{ki}^{t-1}$}
\STATE $g_k^{PC} \gets g_k^{PC} - $ \\ \quad \ $\frac{||g_k^{PC}||(\phi_{ki}^{t-1} \sqrt{1-(\hat{\phi}_{ki}^{t-1})^2}-\hat{\phi}_{ki}^{t-1}\sqrt{1-(\phi_{ki}^{t-1})^2})}{||g_i^t||\sqrt{1-(\hat{\phi}_{ki}^{t-1})^2}} g_i^t$
\ENDIF
\STATE $\hat{\phi}_{ki}^t \gets \delta \hat{\phi}_{ki}^{t-1} + (1-\delta)\phi_{ki}^{t-1}$
\ENDFOR
\ENDFOR
\STATE $g_t^{global} \gets \frac{1}{K} \sum_{k=1}^K g_k^{PC}$
\end{algorithmic}
\end{algorithm}

\section{Experiments}
\subsection{Hyper-parameters}
We perform cross-validation in the training data to find the best hyper-parameters for all methods. We verify all the methods with their hyper-parameters as listed in Table \ref{tabel1}. We take the best performance of each method for the comparison.

\begin{table}[]
    \centering
    \resizebox{\linewidth}{!}{
    \begin{tabular}{l| l}
         \hline
         \hline
         Method &hyper-parameters  \\
         \hline
         FedAvg & $lr \in \{.001, .002, .005, .01, .02\}$ \\
         FedFB & $lr \in \{.001, .005, .01, .02\}$ \\
         FPFL &  $lr \in \{.001, .005, .01, .02\}$, $\gamma \in \{.01, .05, .08\}$ \\
         FairFed & $lr \in \{.001, .005, .01, .02\}$, $\beta \in \{.01, .05, .1, 1, 2, 5\}$ \\
         q-FFL & $lr \in \{.001, .005, .01, .02\}$, $q \in \{.01, .02, 1, 2, 5, 10\}$ \\
         DRFL & $lr \in \{.001, .005, .01, .02\}$, $q \in \{.01, .02, 1, 2, 5, 10\}$ \\
         Ditto & $lr \in \{.001, .005, .01, .02\}$, $\lambda \in \{.01, .05, .1, .5, 1, 5\}$ \\
         FedMGDA+ & $lr \in \{.001, .005, .01, .02\}$ \\
         \hline
         \methodname{} & $\beta \in \{.2, .4, .6, .8, 1\}$, $\delta \in \{.001, .01, .1\}$ \\
         \hline
    \end{tabular}
    }
    \caption{Method specific hyper-parameters. $lr$ is the learning rate of the predictive model, $\gamma$ is the learning rate of the Lagrange multipliers, $\beta$ is the fairness budget in FairFed, $q$ is the aggregated re-weighted parameter, and $\lambda$ controls the interpolation between local and global models. As for \methodname{}, $\beta$ modulates the extent of conflict mitigation, and $\delta$ is the Expomemtial Moving Average (EMA) decay. }
    \label{tabel1}
\end{table}

\subsection{Estimation on Group fairness}
Table \ref{table2} outlines the results with low data heterogeneity among clients, where we randomly assign 30\%, 30\%, 20\%, 10\%, 10\% of the samples from group 0 and 10\%, 20\%, 20\%, 20\%, 30\% of the samples from group 1 to five clients, respectively.

\subsection{Estimation on Client Fairness}
To further solidify \methodname{}'s efficacy, we furnish the number of iterations imperative for all compared methods to attain their optimal performance in Figure \ref{fig1}. This visual depiction affords insights into the convergence trajectories undertaken by distinct methods over iterations. Notably, \methodname{} exhibits commendable performance levels and converges towards the pinnacle of client fairness within a noticeably fewer (or comparable) count of communication rounds. This clearly underscores \methodname{}'s capacity to efficiently train the model, attaining the desired accuracy and client fairness benchmarks with commendable efficacy.

\begin{figure}[!t]
    \centering
    \subfigure[Adult]
    {
    \includegraphics[width=8cm]{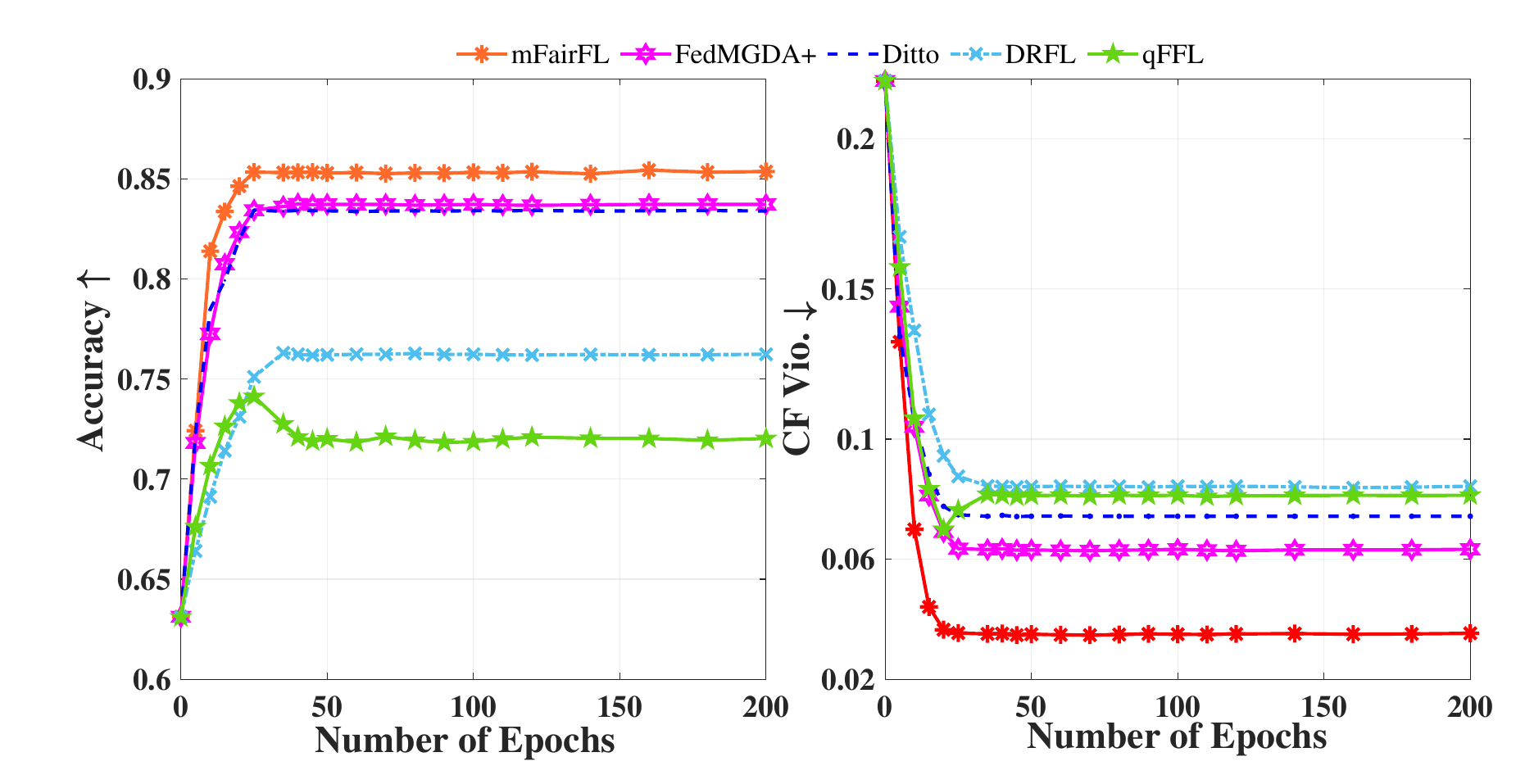}
    \label{adult}}
    \hfil
    \subfigure[Compas]
    {
    \includegraphics[width=8cm]{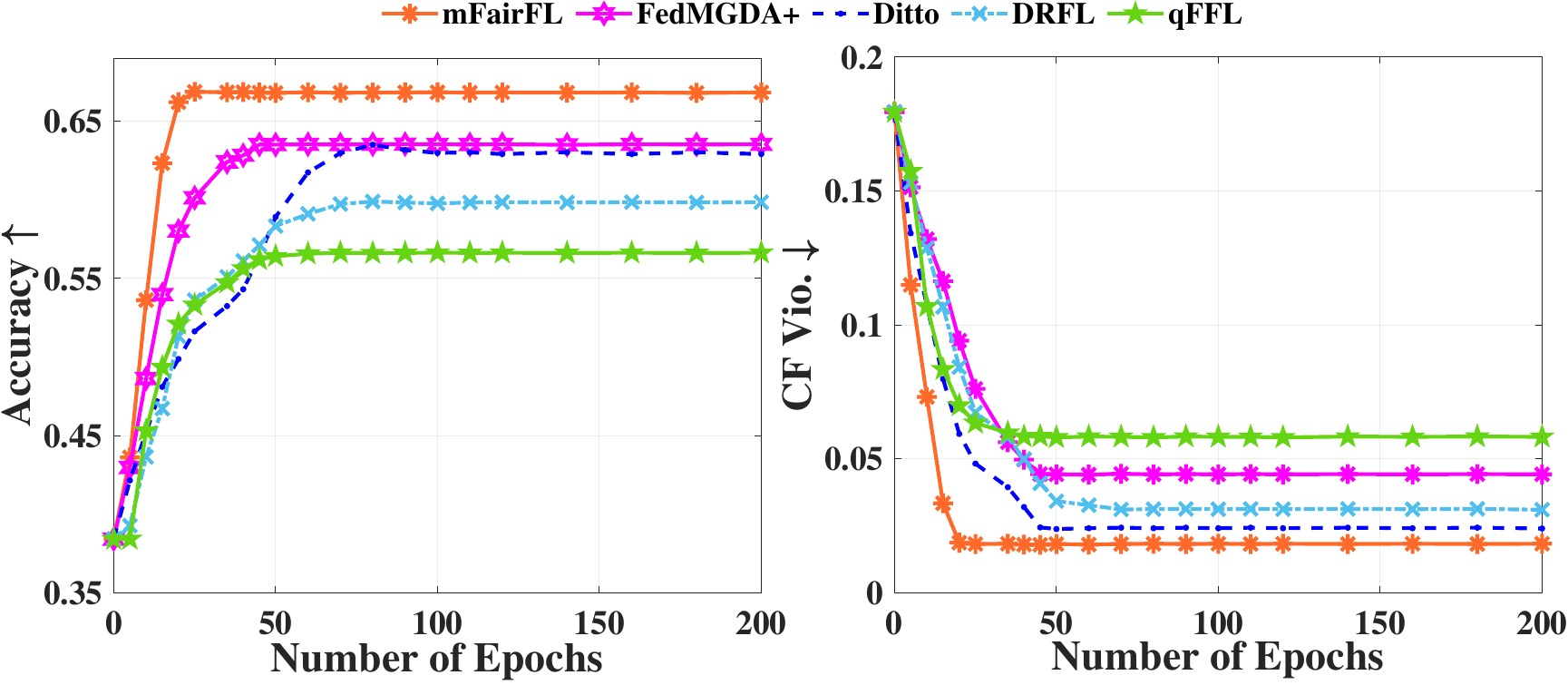}
    \label{dutch}}
    \subfigure[Bank]
    {
    \includegraphics[width=8cm]{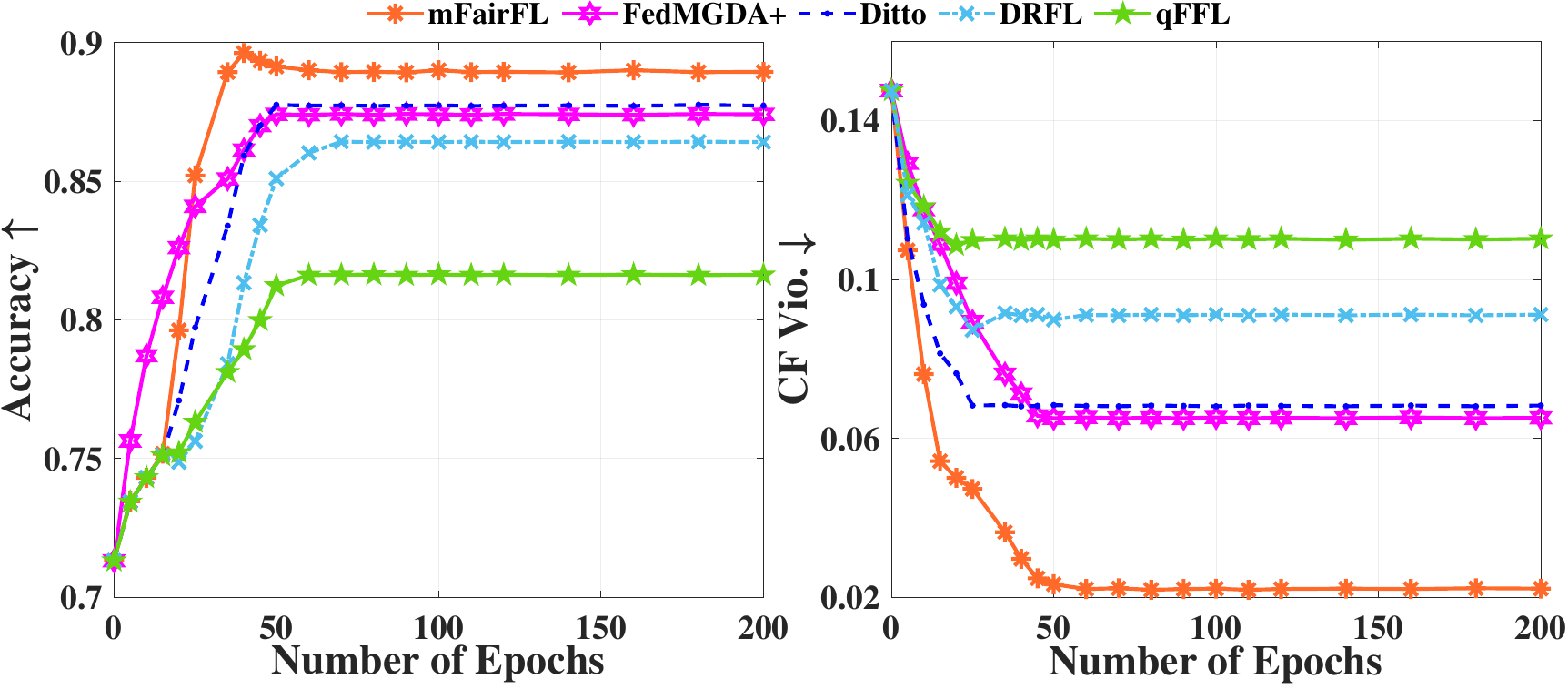}
    \label{dutch}}
    \caption{The performance in term of Accuracy and Client fairness of each method vs. the number of epochs.}
    \label{fig1}
\end{figure}

\begin{table*}[!t]
    \centering
    \resizebox{0.9\linewidth}{!}{
    \renewcommand\arraystretch{1.1}
    \begin{tabular}{r|l l l l l|l l l l l|l l l l l}
        \hline
        \hline
        \multirow{2}{*}{} & \multicolumn{5}{c|}{Adult} & \multicolumn{5}{c|}{Compas} & \multicolumn{5}{c}{Bank} \\
        \cline{3-5}\cline{8-10}\cline{13-15}
        & Acc. & DP & EO & AP & CF & Acc. & DP & EO & AP & CF & Acc. & DP & EO & AP & CF \\
        \hline
        IndFair & .773$\bullet$ & .044$\bullet$ & .058$\bullet$ & .043$\bullet$ & - & .587$\bullet$ & .089$\bullet$ & .097$\bullet$ & .083$\bullet$ & - & .844 & .016$\bullet$ & .014$\bullet$ & .017$\bullet$ & - \\
        FedAvg-F & .739$\bullet$ & .124$\bullet$ & .088$\bullet$ & .132$\bullet$ & .218$\bullet$ & .576$\bullet$ & .057$\bullet$ & .066$\bullet$ & .063$\bullet$ & .192$\bullet$ & .847$\bullet$ & .012$\bullet$ & .016$\bullet$ & .015$\bullet$ & .129$\bullet$ \\
        \hline
        FedFB & .782$\bullet$ & .012 & .009 & \bf .007 & .046$\bullet$ & .565$\bullet$ & .019$\bullet$ & .016$\bullet$ & .018$\bullet$ & .028 & .836 & .009 & .014$\bullet$ & .010 & .048$\bullet$ \\
        FPFL & .766$\bullet$ & .024$\bullet$ & .013$\bullet$ & .021$\bullet$ & .233$\bullet$ & .558$\bullet$ & .027$\bullet$ & .022$\bullet$ & .023$\bullet$ & .183$\bullet$ & .819$\bullet$ & .016$\bullet$ & .007 & .009 & .144$\bullet$ \\
        FairFed & .767$\bullet$ &  .010 & .003 & \bf .007 & .239$\bullet$ & .564$\bullet$ & \bf .007 & \bf .006 & \bf .005 & .194$\bullet$ & .843$\bullet$ & .004 & .003 & .003 & .151$\bullet$ \\
        FedMGDA+ & \bf .845$\circ$ & .305$\bullet$ & .213$\bullet$ & .291$\bullet$ & .046$\bullet$ & \bf .640$\circ$ & .136$\bullet$ & .127$\bullet$ & .142$\bullet$ & .045$\bullet$ & \bf .876$\circ$ & .088$\bullet$ & .086$\bullet$ & .089$\bullet$ & .053$\bullet$ \\
        \hline
        CenFL & .812 & .014 & .008 & .013 & - & .616 & .014$\bullet$ & .008 & .011$\bullet$ & - & .866$\circ$ & \bf .001 & \bf .000 & \bf .002 & - \\
        \hline
        \hline
        \textbf{mFairFL} & .798 & \bf .008 & \bf .002 & .012 & \bf .029 & .605 & .008 & \bf .006 & .009 & \bf .022 & .853 & .003 & .002 & .004 & \bf .026 \\
        \hline
        
    \end{tabular}
    }
    \caption{Accuracy and the violation of \emph{Group fairness} and \emph{Client Fairness} results on the three datasets with low data heterogeneity among clients. The best results in fairness are highlighted in \textbf{boldface}. $\circ/\bullet$  indicates that \methodname{} is statistically worse/better than the compared method by student pairwise $t$-test at 95\% confident level. `-' implies not applicable.}
    \label{table2}
\end{table*}

\begin{table*}[!t]
    \centering
    \resizebox{\linewidth}{!}{
    \renewcommand\arraystretch{1.2}
    \begin{tabular}{r|l l l l l l l l|l l l l l l l l|l l l l l l l l}
        \hline
        \hline
        \multirow{3}{*}{} & \multicolumn{8}{c|}{Adult} & \multicolumn{8}{c|}{Compas} & \multicolumn{8}{c}{Bank} \\
        \cline{3-8}\cline{11-16}\cline{19-24}
        & \multirow{2}*{Acc.} & \multicolumn{2}{c}{DP} & \multicolumn{2}{c}{EO} & \multicolumn{2}{c}{AP} & \multirow{2}*{CF} & \multirow{2}*{Acc.} & \multicolumn{2}{c}{DP} & \multicolumn{2}{c}{EO} & \multicolumn{2}{c}{AP} & \multirow{2}*{CF} & \multirow{2}*{Acc.} & \multicolumn{2}{c}{DP} & \multicolumn{2}{c}{EO} & \multicolumn{2}{c}{AP} & \multirow{2}*{CF} \\
        &  & $S_1$  & $S_2$ & $S_1$  & $S_2$ & $S_1$  & $S_2$ & & & $S_1$  & $S_2$ & $S_1$  & $S_2$ & $S_1$  & $S_2$ & & & $S_1$  & $S_2$ & $S_1$  & $S_2$ & $S_1$  & $S_2$ & \\
        \hline
        IndFair & .754$\bullet$ & .091$\bullet$ & .063$\bullet$ & .066$\bullet$ & .067$\bullet$ & .074$\bullet$ & .057$\bullet$ & - & .545$\bullet$ & .076$\bullet$ & .071$\bullet$ & .088$\bullet$ & .065$\bullet$ & .079$\bullet$ & .067$\bullet$ & - & .817$\bullet$ & .019$\bullet$ & .046$\bullet$ & .022$\bullet$ & .031$\bullet$ & .020$\bullet$ & .033$\bullet$ & -  \\
        FedAvg-F & .704$\bullet$ & .212$\bullet$ & .098$\bullet$ & .137$\bullet$ & .113$\bullet$ & .087$\bullet$ & .198$\bullet$ & .088$\bullet$ & .376$\bullet$ & .538$\bullet$ & .055$\bullet$ & .059$\bullet$ & .067$\bullet$ & .061$\bullet$ & .057$\bullet$ & .062$\bullet$ & .234$\bullet$ & .816$\bullet$ & .022$\bullet$ & .034$\bullet$ & .033$\bullet$ & .036$\bullet$ & .036$\bullet$ & .118$\bullet$  \\
        \hline
        FedFB & .757 & .011 & .010 & .009 & .013 & .009 & .011 & .055$\bullet$ & .546$\bullet$ & .019$\bullet$ & .016$\bullet$ & .016$\bullet$ & .017$\bullet$ & .022$\bullet$ & .018$\bullet$ & .037 & .821 & .008 & .016$\bullet$ & .008 & .016$\bullet$ & .010 & .014 & .062$\bullet$  \\
        FPFL & .742$\bullet$ & .026$\bullet$ & .028$\bullet$ & .019$\bullet$ & .016$\bullet$ & .021$\bullet$ & .027$\bullet$ & .437$\bullet$ & .534$\bullet$ & .036$\bullet$ & .028$\bullet$ & .023$\bullet$ & .022$\bullet$ & .030$\bullet$ & .027$\bullet$ & .244$\bullet$ & .809$\bullet$ & .009$\bullet$ & .020$\bullet$ & .013$\bullet$ & .024$\bullet$ & .012 & .017$\bullet$ & .107$\bullet$  \\
        FairFed & .744$\bullet$ & .011 & \bf .007 & .008 & .010 & .010 & \bf .009 & .311$\bullet$ & .532$\bullet$ & \bf .007 & .008 & \bf .004 & .006 & .005 & .009 & .252$\bullet$ & .804$\bullet$ & .004 & .013 & .007 & .009 & .005 & .009 & .135$\bullet$  \\
        FedMGDA+ & \bf .832$\circ$ & .237$\bullet$ & .216$\bullet$ & .237$\bullet$ & .193$\bullet$ & .236$\bullet$ & .201$\bullet$ & .062$\bullet$ & \bf .634$\circ$ & .136$\bullet$ & .144$\bullet$ & .136$\bullet$ & .132$\bullet$ & .138$\bullet$ & .128$\bullet$ & .044$\bullet$ & \bf .861$\circ$ & .089$\bullet$ & .135$\bullet$ & .081$\bullet$ & .129$\bullet$ & .088$\bullet$ & .133$\bullet$ & .066$\bullet$  \\
        \hline
        CenFL & .783 & .012 & .009 & .010 & \bf .008 & .013 & .010 & - & .587 & .015 & .013 & .011 & .009 & .013 & .011 & - & .848 & \bf .002 & .011 & \bf .003 & \bf .008 & \bf .001 & .009 & -  \\
        \hline
        \hline
        \textbf{mFairFL} & .774 & \bf .010 & .009 & \bf .004 & .011 & \bf .006 & .011 & \bf .038 & .568 & .008 & \bf .004 & .006 & \bf .005 & \bf .003 & \bf .006 & \bf .024 & .832 & .003 & \bf .007 & .004 & .010 & .005 & \bf .008 & \bf .026  \\
        \hline
        
    \end{tabular}
    }
    \caption{Accuracy and the violation of \emph{Group fairness} and \emph{Client Fairness} results on the three datasets with high data heterogeneity among clients. The best results in fairness are highlighted in \textbf{boldface}. $\circ/\bullet$  indicates that \methodname{} is statistically worse/better than the compared method by student pairwise $t$-test at 95\% confident level. `-' implies not applicable.}
    \label{table3}
\end{table*}

\subsection{Experiments on general settings}
In this section, we conduct experiments in more realistic scenarios, where we may encounter multiple domain values or even multiple sensitive attributes. To this end, we treat `gender' (denoted as $S_1$) and `age' (denoted as $S_2$) as the sensitive attributes for Adult dataset. We categorize `age' into three sensitive groups: people aged from 25 to 40, 41 to 65 and the remaining age (treated as disadvantaged group). We treat `gender' (denoted as $S_1$) and `race' (denoted as $S_2$) as the sensitive attributes for COMPAS dataset. We treat `age' (denoted as $S_1$) and `marital' (denoted as $S_2$) as the sensitive attributes for Bank dataset, where the number of sensitive groups for `age' is three, indicating people with age ranging from 20 to 40, 41 to 60 and the remaining ages (treated as disadvantaged group). We also draw samples from each group across five clients at a ratio of 50\%, 10\%, 10\%, 20\%, 10\%, and 10\%, 40\%, 30\%, 10\%, 10\%, respectively, to simulates high data heterogeneity among the clients. The corresponding experimental results are shown in Table \ref{table3}. Our observations from this comparative analysis are as follows: \\
\noindent (i) As expected, \methodname{} significantly improves fairness, achieving a similar level of fairness as CenFL. This not only validates the effectiveness of \methodname{} in adeptly training fair models for sensitive groups in decentralized settings, but also demonstrates its adaptability to more realistic scenarios involving multiple sensitive attributes and values. \\
\noindent (ii) The underperformance of IndFair in terms of group fairness highlights that training fair models exclusively on local data in  decentralized settings with high data heterogeneity falls far short of achieving group fairness at a population level. Besides, the group fairness achieved by FedAvg-f lags behind the fairness demonstrated across the entire population. Moreover, it faces another dimension of fairness issue where the trained global model exhibits bias towards certain clients. These facts highlight the inherent limitations of merely transplanting fairness techniques onto FL. \\
\noindent (iii) \methodname{} demonstrates a superior ability to strike a balance between multi-dimensional fairness, in term of both group fairness and client fairness, and accuracy in scenarios involving multiple sensitive attributes with multiple domain values, surpassing FedFB, FPFL, FairFed, and FedMGDA+. This is because the compared methods either solely focus on group fairness or solely concentrate on client fairness during the global model training. Moreover, they neglect the negative impacts of conflicting gradients with large differences in the magnitudes, resulting in a degradation in the model performance. In contrast,  \methodname{} emphasizes multi-dimensional fairness during model training, and before aggregating the local models, it carefully examines the presence of conflicting gradients. It then adjusts the magnitude and direction of conflicting gradients to mitigate the adverse effects of such conflicts.

\section{Derivation details of Eq. (17)}
As discussed in \textbf{The Aggregation Phase}, the goal of \methodname{} is to adjust the gradients of conflicting clients at each communication round $t$, to align with the currently desired similarity criteria. Suppose at $t$-th communicaiton round, we have two conflicting clients $i$, $j$ with the gradient similarity $\phi_{ij}^t$ (where $\phi_{ij}^t=\cos(g_i^t, g_j^t)=\frac{g_i^t \cdot g_j^t}{||g_i^t|| \cdot ||g_j^t||}$). Then, \methodname{} alters their gradients, so as to match the desired gradient similarity $\hat{\phi}_{ij}^t$. To this end, \methodname{} adjusts the magnitude and orientation of $g_i^t$, denoted as $\hat{g}_i^t = c_1 \cdot g_i^t + c_2 \cdot g_j^t$.  Without loss of generality, we set $c_1=1$ and solve for $c_2$ so that $cos(\hat{g}_i^t, g_j^t) = \frac{\hat{g}_i^t \cdot g_j^t}{||\hat{g}_i^t|| \cdot ||g_j^t||} = \hat{\phi}_{ij}^t$. For the sake of conciseness, we define the angle between $g_i^t$ and $g_j^t$ as $\theta$, and the angle between $\hat{g}_i^t$ and $g_j^t$ as $\beta$. Based on Laws of Sinces, we must have:
\begin{equation}
    \frac{||g_i^t||}{\sin(\beta)} = \frac{c_2||g_j^t||}{\sin(\theta-\beta)}
    \label{eqS1}
\end{equation}
and thus, we can solve for $c_2$ as follows:
\begin{equation}
    \begin{aligned}
        & \ \frac{||g_i^t||}{\sin(\beta)} = \frac{c_2||g_j^t||}{\sin(\theta-\beta)} \\
        & \Rightarrow \frac{||g_i^t||}{\sin(\beta)} = \frac{c_2||g_j^t||}{\sin(\theta)\cos(\beta)-\cos(\theta)\sin(\beta)} \\
        & \Rightarrow \frac{||g_i^t||}{\sqrt{1-(\hat{\phi}_{ij}^t)^2}} = \frac{c_2||g_j^t||}{\hat{\phi}_{ij}^t \sqrt{1-(\phi_{ij}^t)^2}-\phi_{ij}^t \sqrt{1-(\hat{\phi}_{ij}^t)^2}} \\
        & \Rightarrow c_2 = - \frac{||g_i^t||(\phi_{ij}^t \sqrt{1-(\phi_{ij}^t)^2}-\hat{\phi}_{ij}^t \sqrt{1-(\hat{\phi}_{ij}^t)^2})}{||g_j^t||\sqrt{1-(\hat{\phi}_{ij}^t)^2}}
    \end{aligned}
\end{equation}
As a result, we deduce the gradient adjustment rule in Eq. (17), which allows us to achieve arbitrary target similarity value for the gradients of any two clients.

\section{The proof of Theorem 1-3}
Suppose there are $K$ different clients, and each of which has its own dataset $\mathcal{D}_k=\{s_k, \mathbf{x}_k, y_k\}_{i=1}^{n_k}$, where $S$ is the sensitive attribute, $Y$ is the label, and $\mathbf{X}$ is other observational attributes. We assume the data distribution described as follows: let $\mathrm{x|s,k} \sim \mathcal{P}_k^s$ where $\mathcal{P}_k^s$ is the distribution, $\mathrm{s|k} \sim \mathrm{Bern}(q_k)$ where $q_k \in (0,1) for s$, and $\hat{y}|\mathbf{x},s \sim \mathrm{Bern}(f(\mathbf{x},s))$ where $f \in \mathcal{F}$ is the randomized classifier. Table \ref{tabel4} summarize the glossary of commonly used symbols.

\begin{table}[]
    \centering
    \resizebox{\linewidth}{!}{
    \begin{tabular}{l l| l l}
         \hline
         \hline
         Symbol &Meaning &Symbol &Meaning  \\
         \hline
         $\mathbf{x}$ & Non-sensitive attributes & $f$ & Randomised classifier \\
         s & Sensitive attribute & $\eta(\mathbf{x})$ & $\mathbb{P}(y=1|\mathbf{X}=\mathbf{x})$ \\
         $y$ & Label & $\mathcal{P}_k^s$ & Distribution \\
         $\hat{y}$ & Prediction & MD & Mean difference \\
         $\alpha_k$ & Bias tolerance of client $k$ & $q$ & $s \sim \mathrm{Bern}(q)$ \\
         \hline
    \end{tabular}
    }
    \caption{The glossary of commonly used symbols required for the proof of Theorem 1.}
    \label{tabel4}
\end{table}

\subsection{The proof of Theorem 1}
\begin{theorem}[Necessity for FL]
If the data distribution is highly heterogeneous across  clients, then $\min FN(g_{\alpha}^{Ind}) > \min FN(g_{\alpha}^{Fed})$.
\label{theo1}
\end{theorem}

\textbf{Proof.} To prove Theorem 1, we firstly need to solve for the minimum achievable fairness violation scores of \emph{IndFair} and \emph{FedFair}, respectively.

\emph{IndFair:} each client $k$ trains a fairness model independently. Our prove for \emph{IndFair} is similar to \cite{li2019convergence}. Suppose $p_k^s(\cdot)$ is the probability density function of $\mathcal{P}_k^s$ where $s=s^+,s^-$, and $p_k^{\mathbf{x},s}(\mathbf{x},s)$ is the joint distribution of $\mathbf{x}$ and $s$. The corresponding error rate for solving such model can be defined as follows:
\begin{equation}
    \begin{aligned}
        & \mathbb{P}(\hat{y} \neq y) \\
        = & \int_{\mathbf{X}} \sum_{s \in S}[f_k(\mathbf{x},s)(1-\eta(\mathbf{x}))+(1-f_k(\mathbf{x},s))\eta(\mathbf{x})] \\
        & \quad p_k^{\mathbf{x},s}(\mathbf{x},s) dx \\
        = & \mathbb{E}_{\mathbf{x},s}f_k(\mathbf{x},s)(1-2\eta(\mathbf{x}))+\mathbb{P}(y=1)
    \end{aligned}
\end{equation}
and the corresponding fairness constraint is defined as follows:
\begin{equation}
    \begin{aligned}
        & \mathbb{P}(\hat{y}=1|s=s^-) - \mathbb{P}(\hat{y}=1|s=s^+) \\
        = & \int_{\mathbf{X}} f_k(\mathbf{x},s^-)p_k^{s^-}dx - \int_{\mathbf{X}} f_k(\mathbf{x},s^+)p_k^{s^+}dx \\
        = & \int_{\mathbf{X}} \sum_{s \in S} \mathbb{I}[s=s^-] f_k(\mathbf{x},s^-) \frac{p_k^{\mathbf{x},s}(\mathbf{x},s)}{\mathbb{P}(s=s^-)}dx \\
        &  - \int_{\mathbf{X}} \sum_{s \in S} \mathbb{I}[s=s^+] f_k(\mathbf{x},s^+) \frac{p_k^{\mathbf{x},s}(\mathbf{x},s)}{\mathbb{P}(s=s^+)}dx \\
        = & \mathbb{E}_{\mathbf{x},s}[f_k(\mathbf{x},s^-)\frac{\mathbb{I}[s=s^-]}{1-q_k}-f_k(\mathbf{x},s^+)\frac{\mathbb{I}[s=s^+]}{q_k}]
    \end{aligned}
\end{equation}
where $\mathbb{I}[a]$ is the indicator function. $\mathbb{I}[a]=1$, if $a$ is true; $\mathbb{I}[a]=0$, otherwise. Therefore, the objective function in Eq. (7) can be rewritten as follows:
\begin{subequations} \label{eqS5}
    \begin{align}
        \min_{f_k \in \mathcal{F}} & \quad \mathbb{E}_{\mathbf{x},s}f_k(\mathbf{x},s)(1-2\eta(\mathbf{x}))+\mathbb{P}(y=1) \tag{S5a}\label{S5a} \\
        s.t. & \quad |\mathbb{E}_{\mathbf{x},s}[f_k(\mathbf{x},s^-)\frac{\mathbb{I}[s=s^-]}{1-q_k} \notag \\
        & \quad -f_k(\mathbf{x},s^+)\frac{\mathbb{I}[s=s^+]}{q_k}]| = \alpha_k \tag{S5b}\label{S5b}
    \end{align}
\end{subequations}
We denote the function that minimizes the error rate (ERM) in Eq. \eqref{S5a} as $\tilde{f}\in \mathcal{F}$, which is easy to see that:
\begin{equation}
    \tilde{f}_k(\mathbf{x}) \in \{ \mathbb{I}[\eta(\mathbf{x})>\frac{1}{2}]+\beta\mathbb{I}[\eta(\mathbf{x})=\frac{1}{2}]: \beta \in [0,1] \}
    \label{eqS6}
\end{equation}

Furthermore, by introducing Lagrange multipliers $\lambda$, Eq. \eqref{eqS5} is also equivalent to 
\begin{equation}
    \begin{aligned}
        \min_{f_k \in \mathcal{F}} & \quad  \mathbb{E}_{\mathbf{x},s}f_k(\mathbf{x},s)(1-2\eta(\mathbf{x})) \\
        & - \lambda \mathbb{E}_{\mathbf{x},s} \big( f_k(\mathbf{x},s^-)\frac{\mathbb{I}[s=s^-]}{1-q_k}-f_k(\mathbf{x},s^+)\frac{\mathbb{I}[s=s^+]}{q_k} \big)
    \end{aligned}
    \label{eqS7}
\end{equation}
where Eq. \eqref{eqS7} can be rewritten as follows:
\begin{equation}
    \min_{f_k \in \mathcal{F}} \quad  \mathbb{E}_{\mathbf{x},s}f_k(\mathbf{x},s)\big( 1-2\eta(\mathbf{x}) - \lambda \frac{\mathbb{I}[s=s^-]}{1-q_k} + \lambda \frac{\mathbb{I}[s=s^+]}{q_k} \big)
    \label{eqS8}
\end{equation}
Then, the goal becomes to select a suitable $\lambda$ such that the fairness constrained optimization problem in Eq. \eqref{eqS5} becomes an unconstrained problem.

According to Eq. \eqref{eqS6}, the solution of Eq. \eqref{eqS8} is as follows:
\begin{equation}
    \tilde{f}_k(\mathbf{x}) \in \{ \mathbb{I}[s_k(\mathbf{x},s)>0]+\beta\mathbb{I}[s_k(\mathbf{x},s)=0]: \beta \in [0,1] \}
    \label{eqS9}
\end{equation}
where 
\begin{equation}
    s_k(\mathbf{x},s)=
    \begin{cases}
    -1+2\eta(\mathbf{x})+\frac{\lambda_{\alpha_k}}{1-q_k}& s=s^- \\
    -1+2\eta(\mathbf{x})-\frac{\lambda_{\alpha_k}}{q_k}& s=s^+
    \end{cases}
\end{equation}
Then, our goal is to find the value of Lagrange multipliers $\lambda \in [-\max(q_k,1-q_k),\max(q_k,1-q_k)]$ such that $f_k$ satisfies the fairness constraint in Eq. \eqref{S5b}.

Consider the mean difference between the positive rate of two sensitive groups:
\begin{equation}
    \begin{aligned}
        MD(\tilde{f}_k) =& \mathbb{P}(\hat{y}=1|s=s^-,i=k) - \mathbb{P}(\hat{y}=1|s=s^+,i=k) \\
        =& \int_{-\infty}^{+\infty} \mathbb{I}[\eta(\mathbf{x})>\frac{1}{2}-\frac{\lambda}{2(1-q)}]d\mathcal{P}_k^{s^-} \\
        & - \int_{-\infty}^{+\infty} \mathbb{I}[\eta(\mathbf{x})>\frac{1}{2}+\frac{\lambda}{q}]d\mathcal{P}_k^{s^+} \\
        =& \int_{\eta^{-1}(\frac{1}{2}-\frac{\lambda}{2(1-q)})}^{+\infty}d\mathcal{P}_k^{s^-} - \int_{\eta^{-1}(\frac{1}{2}+\frac{\lambda}{q})}^{+\infty}d\mathcal{P}_k^{s^+} \\
        =& g_k(\lambda)
    \end{aligned}
    \label{eqS11}
\end{equation}
where $g_k(\cdot): [-\max(q_k,1-q_k),\max(q_k,1-q_k)] \to [-1,1]$ is a strictly monotone increasing function, which is the mean difference on $k$-th client (i.e., $\mathbb{E}_{\mathbf{x} \sim \mathcal{P}_k^{s^-}}f(\mathbf{x},s^-)-\mathbb{E}_{\mathbf{x} \sim \mathcal{P}_k^{s^+}}f(\mathbf{x},s^+)$). Therefore, if and only if $\lambda_{\alpha_k}=g^{-1}(\alpha_k)$, $\tilde{f}$ meets fairness requirement of Eq. \eqref{S5b}.

Specifically, let $\beta=0$, then the solution of \emph{IndFair} can be defined as follows:
\begin{equation}
    f_k^{\alpha_k}(\mathbf{x},s)=
    \begin{cases}
    \mathbb{I}[\eta(\mathbf{x}>\frac{1}{2}-\frac{\lambda_{\alpha_k}}{2(1-q_k)})]& s=s^- \\
    \mathbb{I}[\eta(\mathbf{x})>\frac{1}{2}+\frac{\lambda_{\alpha_k}}{q_k}]& s=s^+
    \end{cases}
    \label{eqS12}
\end{equation}
where we denote $\lambda_{\alpha_k}$ as
\begin{equation}
    \lambda_{\alpha_k} = g_k^{-1}(sign(g_k(0))\min(\alpha_k,|g_k(0)|))
    \label{eqS13}
\end{equation}
As such, $\lambda_{\alpha_k}$ has the following useful properies, which is proved in \textbf{Lemma 7} by \citet{zeng2021improving}:
\begin{itemize}
 \item If $\alpha_k < |g_k(0)|$, then $MD(f_k^\alpha)=\lambda_{\alpha_k}\neq0$, and $g_k(\lambda_{\alpha_k})=\mathrm{sign}(g_k(0))\alpha_k$.
 \item If $\alpha_k \geq |g_k(0)|$, then $\lambda_{\alpha_k}=0$ and $MD(f_k^\alpha)=g_k(\lambda_{\alpha_k})=g_k(0)$.
 \item If $g_k(0)>0$ or $g_k(0)<0$, then for any $\alpha_k\geq0$, we have $\lambda_{\alpha_k}\leq0$ or $\lambda_{\alpha_k}\geq0$, respectively.
 \end{itemize}

To proof Theorem \ref{theo1}, we firstly need to show the limitation of \emph{IndFair}, i.e., \emph{IndFair} cannot achieve fairness violation scores of zero, which is verified as the following Lemma:
\begin{lemma}
    If there exists at least a pair of clients $n$ and $m$ where $g_n(0)\cdot g_m(0)<0$ and $\rho(\alpha_n,\alpha_m)(g_n(0)+g_m(0))>0$ for all $\alpha_n,\alpha_m \in [0,c]$, where $\rho(\alpha_n,\alpha_m)=g_n(g_m^{-1}(\mathrm{sign}(g_m(0))\alpha_m))+g_j(g_n^{-1}(\mathrm{sign}(g_n(0))\alpha_n))+\mathrm{sign}(g_n(0))\alpha_n + \mathrm{sign}(g_m(0))\alpha_m$, and $c=\min\{|g_n(0)|,|g_m(0)|\}$; then $FN(g_{\alpha}^{Ind})\geq \delta=\min\{\rho(\alpha_1,\cdots,\alpha_K):\alpha_1,\cdots,\alpha_K \in [0,c]\}$.
\label{lemma1}
\end{lemma}
\textbf{Proof.} We denote mean difference between two sensitive groups for clients $i$ and $j$ as
\begin{equation}
    \begin{aligned}
        MD(f_{n,m})&=\sum_{p \in \{n,m\}}[\mathbb{P}(\hat{y}=1|s^-,i=p) \\
        & \quad -\mathbb{P}(\hat{y}=1|s^+,i=p)] \\
        &=\sum_{p,q \in \{n,m\}}[\mathbb{E}_{x \sim \mathcal{P}_p^{s^-}}f_q^{\alpha_q}(x,s^-) \\
        & \quad - \mathbb{E}_{x \sim \mathcal{P}_p^{s^+}}f_q^{\alpha_q}(x,s^+) ] \\
        &= g_n(\lambda_{\alpha_n})+g_n(\lambda_{\alpha_m})+g_m(\lambda_{\alpha_n})+g_m(\lambda_{\alpha_m})
    \end{aligned}
\end{equation}
Without loss of generality, we assume $|g_n(0)|<|g_m(0)|$. We consider $g_m(0)>0$. (The discussion $g_m(0)<0$ is similar to $g_m(0)>0$ case.) By $g_n(0)\cdot g_m(0)<0$ and $\rho(\alpha_n,\alpha_m)(g_n(0)+g_m(0))>0$, we have $g_n(0)<0$ and $\rho(\alpha_n,\alpha_m)>0$.

\emph{Case 1.} $\alpha_n > |g_n(0)|$, and $\alpha_m > |g_m(0)|$, which represents ERM is fair on both clients $n$ and $m$. 

By Eq. \eqref{eqS13}, we have $\lambda_{\alpha_n}=\lambda_{\alpha_m}=0$. Since $g(\cdot)$ is a monotone increasing function, we combine $g_m(0)>0$ and the properies of $\lambda_{\alpha_k}$ to have $g_n(g_m^{-1}(0))<g_n(0)<0$. Then, we apply the above conclusion to obtain:
\begin{equation}
    \begin{aligned}
        MD(f_{n,m})&=2g_n(0)+2g_m(0) \\
        &> 2(g_n(0)+g_m(0)+g_n(g_m^{-1}(0))) \\
        &= 2 \rho(g_n(0),0) \geq \delta
    \end{aligned}
\end{equation}

\emph{Case 2.} $\alpha_n \leq |g_n(0)|$, and $\alpha_m > |g_m(0)|$, which represents ERM is fair on client $m$, but unfair on client $n$.

By Eq. \eqref{eqS13}, we have $\lambda_{\alpha_m}=0$. Since $g(\cdot)$ is a monotone increasing function, we have $\lambda_{\alpha_n}=g_n^{-1}(-\alpha_n)>g_n^{-1}(g_n(0))=0$. Then, applying the above conclusion yields:
\begin{equation}
    \begin{aligned}
        MD(f_{n,m})&=-\alpha_n + g_m(0) + g_n(0) + g_m(\lambda_{\alpha_n}) \\
        &> 2g_n(0)+2g_m(0) \\
        &> 2 \rho(g_n(0),0) \geq \delta
    \end{aligned}
\end{equation}

\emph{Case 3.} $\alpha_n > |g_n(0)|$, and $\alpha_m \leq |g_m(0)|$, which represents ERM is fair on client $n$, but unfair on client $m$.

By Eq. \eqref{eqS13}, we have $\lambda_{\alpha_n}=0$, $\lambda_{\alpha_m}=g_m^{-1}(\alpha_m)>g_m^{-1}(0)$. Then we have:
\begin{equation}
    \begin{aligned}
        MD(f_{n,m})&=g_n(0) + g_n(\lambda_{\alpha_m}) + g_m(0) + \alpha_m) \\
        &> g_n(0) + g_n(g_m^{-1}(0)) + g_m(0) \\
        &= \rho(g_n(0),0) \geq \delta
    \end{aligned}
\end{equation}

\emph{Case 4.} $\alpha_n \leq |g_n(0)|$, and $\alpha_m \leq |g_m(0)|$, which represents ERM is unfair on both clients $n$ and $m$.

By Eq. \eqref{eqS13}, we have $\lambda_{\alpha_n}=g_n^{-1}(\alpha_n)$, $\lambda_{\alpha_m}=g_m^{-1}(\alpha_m)$. Then we can obtain:
\begin{equation}
    \begin{aligned}
        MD(f_{n,m})&=-\alpha_n + \alpha_m + g_n(g_m^{-1}(\alpha_m)) + g_m(g_n^{-1}(-\alpha_n))) \\
        &\leq -\alpha_n + \alpha_n + g_n(g_m^{-1}(\alpha_n)) + g_m(g_n^{-1}(-\alpha_n))) \\
        &= \rho(\alpha_n, \alpha_n) \geq \delta
    \end{aligned}
\end{equation}
Combining all the cases above, we conclude that $MD(f_{n,m}) \geq \min\{\rho(\alpha_n,\alpha_m):\alpha_n,\alpha_m \in [0,c]\}$. Then, combining the rest of the clients, we have:
\begin{equation}
    \begin{aligned}
        &FN(g_{\alpha}^{Ind})=\frac{1}{K^2}|MD(f_{n,m}) \\
        &+\sum_{i,j\neq\{n,m\}}^K[\mathbb{E}_{x \sim \mathcal{P}_i^{s^-}}f_j^{\alpha_q}(x,s^-) - \mathbb{E}_{x \sim \mathcal{P}_i^{s^+}}f_j^{\alpha_q}(x,s^+)]| \\
        &=\frac{1}{K^2} \sum_{i,j=0}^K g_i(\lambda_{\alpha_j}) \\
        &\geq \frac{1}{K^2}(\rho(\alpha_n,\alpha_m) + \sum_{i\neq\{n,m\}}^K\sum_{j\neq i}^K g_i(g_j^{-1}(\mathrm{sign}(g_j(0))\alpha_j)) \\
        &+ \sum_{i\neq\{n,m\}}^K \mathrm{sign}(g_i(0))\alpha_i) \\
        &\geq \min\{\rho(\alpha_1,\cdots,\alpha_K):\alpha_1,\cdots,\alpha_K \in [0,c]\} = \delta
    \end{aligned}
    \label{eqS19}
\end{equation}
Then we finally proof the Lemma \ref{lemma1}.

\emph{FedFair:} For any $\alpha \in [0,1]$, we denote $\alpha_1,\cdots,\alpha_K=\alpha$. Then the global fairness violation is as follows:
\begin{equation}
    \begin{aligned}
        &FN(g_{\alpha}^{Fed})= |\mathbb{E}_{x|s=s^-}f(x,s^-)- \mathbb{E}_{x|s=s^+}f(x,s^+)| \\
        &= |\frac{1}{K} (\sum_{i=1}^K \int_{\mathbf{X}}f(\mathbf{x},s^-) d\mathcal{P}_i^{s^-} - \sum_{i=1}^K \int_{\mathbf{X}}f(\mathbf{x},s^+) d\mathcal{P}_i^{s^+})| \\
        &= \frac{1}{K} |\sum_{i=1}^K MD_i(g_{\alpha}^{Fed})| \\
        & \leq \frac{1}{K} \sum_{i=1}^K |MD_i(g_{\alpha}^{Fed})| = \frac{\alpha_1+\cdots+\alpha_K}{K} = \alpha 
    \end{aligned}
    \label{eqS20}
\end{equation}
Therefore, according to Eq. \eqref{eqS19} and Eq. \eqref{eqS20}, we finally proof that $\min FN(g_{\alpha}^{Ind}) > \min FN(g_{\alpha}^{Fed})$.

\subsection{The proof of Theorem 2}
\begin{theorem}
Suppose there is a set of gradients $G=\{g_1,g_2,...,g_K\}$ where $g_i$ always conflicts with $g_j^{t_j}$ before adjusting $g^{t_j}_j$ to match similarity goal between $g^{t_j}_j$ and $g_i$ ($g^{t_j}_j$ represents the gradient adjusting $g_j$ with the target gradients in $G$ for $t_j$ times). Suppose $\epsilon_1 \leq |cos(g^{t_i}_i, g^{t_j}_j)| \leq \epsilon_2$, $0$$<$$\epsilon_1$$<$$\hat{\phi}_{ij}$$\leq$$\epsilon_2$$\leq 1$, for each $g_i \in G$, as long as we iteratively project $g_i$ onto $g_k$'s normal plane (skipping $g^i$ itself) in the ascending order of $k$=$1,2,\cdots,K$, the larger the $k$ is, the smaller the upper bound of conflicts between the aggregation gradient of global model $g^{global}$ and $g_k$ is. The maximum value of $|g^{global} \cdot g_k|$ is bounded by $\frac{K-1}{K}(\max_i||g_i||)^2\frac{\epsilon_2X_{\max}(1-X_{\min})(1-(1-X_{min})^{K-k})}{X_{\min}}$, where $X_{\max}$=$\frac{\epsilon_2\sqrt{1-\hat{\phi}^2}-\hat{\phi}\sqrt{1-\epsilon_2^2}}{\sqrt{1-\hat{\phi}^2}}$ and $X_{\min}$=$\frac{\epsilon_1\sqrt{1-\hat{\phi}^2}-\hat{\phi}\sqrt{1-\epsilon_1^2}}{\sqrt{1-\hat{\phi}^2}}$.
\label{theo2}
\end{theorem}

\textbf{Proof.} for each gradient $g_i \in G$, we adjust the  magnitude and orientation of $g_i$ to align with the desired similarity criteria between $g_i$ and the target gradient $g_k$ in an increasing order with $k$. Then, we have the following update rules:
\begin{equation}
    \begin{cases}
    g_i^0=g_i& k=0 \\
    g_i^k=g_i^{k-1}- \frac{||g_i^{k-1}||X_{ik}}{||g_k||} \cdot g_k& k=1,...,K;k\neq i \\
    g_i^k = g_i^{k-1}& k=i
    \end{cases}
\end{equation}
where $X_{ik}=\frac{\phi_{ik} \sqrt{1-(\hat{\phi}_{ik})^2}-\hat{\phi}_{ik}\sqrt{1-(\phi_{ik})^2}}{\sqrt{1-(\hat{\phi}_{ik})^2}}$, $\phi_{ik}=\cos(g_i,g_k)$, and $\hat{\phi}_{ik}$ is the gradient similarity goal between $i$ and $k$.

As a result, $g_i^k$ never conflicts with $g_k$, since all potential conflicts have been eliminated by the update rules. Now, we focus on how much the final gradient $g^{global}$ conflicts with each gradient in $G$. Firstly, the last in the reference projection order is $g_K$, and the final global gradient $g^{global}$ will not conflict with it. Then, we focus on the last but second gradient $g_{K-1}$. According to the update rules, we obtain:
\begin{equation}
    g_i^K=g_i^{K-1} - \frac{||g_i^{K-1}||X_{iK}}{||g_K||} \cdot g_K
\end{equation}

Then the global aggregation gradients is:
\begin{equation}
    \begin{aligned}
         & g^{global}=\frac{1}{K}\sum_i^K g_i^K \\
         & = \frac{1}{K} (\sum_{i\neq K}^K (g_i^{K-1}-||g_i^{K-1}|| X_{iK} \cdot \frac{g_K}{||g_K||})+g_K^{K-1})\\
         &= \frac{1}{K} \sum_i^K g_i^{K-1}-\frac{1}{K}\sum_{i\neq K}^K ||g_i^{K-1}|| X_{iK} \frac{g_K}{||g_K||}
    \end{aligned}
    \label{eqS23}
\end{equation}
Since $g_k \cdot \sum_i^K g_i^k \geq 0$, we can compute the conflicts between any gradient $g_k \in G$ and $g^{global}$ by removing the $g_i^k$ from $g^{global}$:
\begin{equation}
    \begin{aligned}
         g_k \cdot g^{global} &\geq g_k \cdot \sum_{j=k}^{K-1}(\frac{-1}{m}\sum_{i \neq j+1}^K||g_i^j||X_{i,j+1} \frac{g_{j+1}}{||g_{j+1}||}) \\
         & = -\frac{||g_k||}{K} \sum_{j=k}^{K-1} \sum_{i \neq j+1}^K ||g_i^j|| X_{i,j+1} \phi_{k,j+1} \\
         & \geq -\frac{\epsilon_2}{K} ||g_k|| \sum_{j=k}^{K-1} \sum_{i \neq j+1}^K ||g_i^j|| X_{\max}
    \end{aligned}
    \label{eqS23}
\end{equation}
where $X_{\max}$=$\frac{\epsilon_2\sqrt{1-\hat{\phi}^2}-\hat{\phi}\sqrt{1-\epsilon_2^2}}{\sqrt{1-\hat{\phi}^2}}$.
\begin{equation}
    |e_k \cdot g^{global}| = |\frac{g_k}{||g_k||} \cdot g^{global}| \leq \frac{\epsilon_2}{K} \sum_{j=k}^{K-1} \sum_{i \neq j+1}^K ||g_i^j|| X_{\max}
    \label{eqS25}
\end{equation}
Therefore, the later $g_k$ serves as the reference target of others, the smaller the upper bound of conflicts between $g^{global}$ and $g_k$ is, since $\sum_{j=k}^{K-1} \sum_{i \neq j+1}^K ||g_i^j|| \leq \sum_{j=k-1}^{K-1} \sum_{i \neq j+1}^K ||g_i^j||$.

According to Eq. \eqref{eqS25}, the upper bound of the conflict between any client gradient $g_k$ and the final global gradient $g^{global}$ is defined as follows:
\begin{equation}
    |g_k \cdot g^{global}| \leq \frac{\epsilon_2}{K} ||g_k|| \sum_{j=k}^{K-1} \sum_{i \neq j+1}^K ||g_i^j|| X_{\max}
    \label{eqS26}
\end{equation}

With the update rule of \methodname{}, we can infer that
\begin{equation}
    \begin{aligned}
         ||g_k^i||^2 &= ||g_k^{i-1}-\frac{||g_k^{i-1}||X_{k,i}}{||g_i||} \cdot g_i||^2 \\
         &= ||g_k^{i-1}||^2 +2||g_k^{i-1}||^2 X_{ki} \phi_{ki} + ||g_k^{i-1}||^2 X_{ki}^2 \\
         &= ||g_k^{i-1}||^2(X_{ki}^2 + 2X_{ki}\phi_{ki}+1) \\
         & \leq ||g_k^{i-1}||^2 (X_{ki}+1)^2
    \end{aligned}
    \label{eqS23}
\end{equation}
Therefore, the maximum value of gradient conflict is bounded by
\begin{equation}
    \begin{aligned}
         |g_k \cdot & g^{global}| \leq \frac{\epsilon_2}{K}||g_k|| \sum_{j=k}^{K-1} \sum_{i \neq j+1}^K ||g_i^j|| X_{\max} \\
         &\leq \frac{\epsilon_2}{K} (\max_{i}||g_i||) \sum_{j=k}^{K-1} \sum_{i \neq j+1}^K ||g_i^0|| X_{\max} (1-X_{\min})^j \\
         &= \frac{K-1}{K} (\max_{i}||g_i||)^2 \epsilon_2 X_{\max} \sum_{j=k}^{K-1} (1-X_{\min})^j \\
         &= \frac{K-1}{K}(\max_i||g_i||)^2 \\
         & \quad \quad \cdot \frac{\epsilon_2X_{\max}(1-X_{\min})(1-(1-X_{min})^{K-k})}{X_{\min}}
    \end{aligned}
    \label{eqS23}
\end{equation}

\subsection{The proof of Theorem 3}
\begin{theorem}
Suppose there are $K$ objective functions $J_1(w), J_2(w),\cdots, J_K(w)$, and each objective function is differentiable and L-smooth. Then \methodname{} will converage to the optimal $w^*$ within a finite number of steps.
\label{theo3}
\end{theorem}

\textbf{Proof.} \emph{Case 1.} If there is no conflict between clients, which indicate that $g_i \cdot g_j \geq \hat{\phi}_{ij}$ for any two clients $i$ and $j$, where $\hat{\phi}_{ij}$ is the similarity goal between $g_i$ and $g_j$. In this case, \methodname{} updates as FedAvg does, simply computing an average of the two gradients and adding it to the parameters to obtain the new parameters $w^{t+1}=w^t-\gamma\bar{g}$, where $\bar{g}=\frac{1}{K}\sum_{i=1}^K g_i$. Thus, when using step size $\gamma \leq \frac{1}{L}$ (where $L$ is the constant of Lipschitz continuous), it will decrease the objective function $\mathcal{L}(w)$ \cite{li2019convergence}. 

\emph{Case 2.} On the other hand, if there exists conflicting gradients for clients, \methodname{} systematically adjusts the magnitude and direction of the client gradient to mitigate gradient conflicts. For the purpose of clarity, suppose there are two clients $i$ and $j$ with gradient conflicts, and their loss are $\mathcal{L}_i$ and $\mathcal{L}_j$, in which $\mathcal{L}_i$ and $\mathcal{L}_j$ are differentiable, and the gradient of them $g_i=\nabla \mathcal{L}_i$ and $g_j=\nabla \mathcal{L}_j$ are Lipschitz continuous with constant $L > 0$. Then, we have $g=g_i+g_j$ as the original gradient and $\tilde{g}=g+c\frac{||g_j||}{||g_i||}g_i+c\frac{||g_i||}{||g_j||}g_j$ as the altered gradient by \methodname{} update rules, such that:
\begin{equation}
    c = \frac{\sin(\varphi_{ij})\cos(\hat{\varphi}_{ij})-\cos(\varphi_{ij})\sin(\hat{\varphi}_{ij})}{\sin(\hat{\varphi}_{ij})}
\end{equation}
where $\varphi_{ij}$ is the angle between $g_i$ and $g_j$, and $\hat{\varphi}_{ij}$ is the disired angle between $g_i$ and $g_j$. $\hat{\varphi}_{ij}\geq \varphi_{ij}$, and $c \geq 0$, since we only consider the angle between two gradients in the range of 0 to $\pi$.

Thus, we obtain the quadratic expansion of $\mathcal{L}$ as
\begin{equation}
    \begin{aligned}
         \mathcal{L}(w^{t+1}) \leq  \mathcal{L}(w^{t}) &+ \nabla \mathcal{L}(w)^T(w^{t+1}-w^t) \\
         & + \frac{1}{2}\nabla^2 \mathcal{L}(w^{t}) ||w^{t+1}-w^{t}||^2
    \end{aligned}
    \label{eqS30}
\end{equation}
and utilize the assumption that $\nabla \mathcal{L}$ is  Lipschitz continuous with constant $L$, Eq. \eqref{eqS30} can be rewritten as follows:
\begin{equation}
    \mathcal{L}(w^{t+1}) \leq \mathcal{L}(w^{t}) + \nabla \mathcal{L}(w^{t})(w^{t+1}-w^t) + \frac{1}{2} L||w^{t+1}-w^{t}||^2
    \label{eqS31}
\end{equation}

Then, we plus in the update rule of \methodname{} to obtain
\begin{equation}
    \begin{aligned}
        \mathcal{L}(w^{t+1}) \leq & \mathcal{L}(w^t) -\gamma \cdot g^T(g+c\frac{||g_j||}{||g_i||}g_i+c\frac{||g_i||}{||g_j||}g_j) \\
        &+ \frac{1}{2}L\gamma^2||g+c\frac{||g_j||}{||g_i||}g_i+c\frac{||g_i||}{||g_j||}g_j||^2 \\
        =& \mathcal{L}(w^t) - (\gamma-\frac{1+c^2}{2}L\gamma^2 \\
        &+ c\phi_{ij}(\gamma-L\gamma^2))(||g_i||^2+||g_j||^2) \\
        &- (2c(\gamma-L\gamma^2)+\phi_{ij}(2\gamma-L\gamma^2(1+c^2))) \\
        &\quad \ (||g_i||\cdot ||g_j||) \\
        =& \mathcal{L}(w^t) - (\gamma-\frac{1+c^2}{2}L\gamma^2)(||g_i||^2+||g_j||^2) \\
        & - 2\phi_{ij}(\gamma-\frac{1+c^2}{2}L\gamma^2)(||g_i||\cdot ||g_j||) \\
        & - (c\phi_{ij}(\gamma-L\gamma^2))(||g_i||^2+||g_j||^2) \\
        & - (2c(\gamma-L\gamma^2))(||g_i||\cdot ||g_j||) \\
        \leq & \mathcal{L}(w^t) - (\gamma-\frac{1+c^2}{2}L\gamma^2)(||g_i||^2+||g_j||^2) \\
        & - 2\phi_{ij}(\gamma-\frac{1+c^2}{2}L\gamma^2)(||g_i||\cdot ||g_j||) \\
        =& \mathcal{L}(w^t) - (\gamma-\frac{1+c^2}{2}L\gamma^2) \\
        &\quad \quad (||g_i||^2+||g_j||^2+2\phi_{ij}||g_i||\cdot||g_j||) \\
        =& \mathcal{L}(w^t) - (\gamma-\frac{1+c^2}{2}L\gamma^2) \\
        &\quad \quad (||g_i||^2+||g_j||^2+2 g_i \cdot g_j) \\
        =& \mathcal{L}(w^t) - (\gamma-\frac{1+c^2}{2}L\gamma^2) ||g_i+g_j||^2 \\
        =& \mathcal{L}(w^t) - (\gamma-\frac{1+c^2}{2}L\gamma^2) ||g||^2
    \end{aligned}
    \label{eqS32}
\end{equation}
The last line implies that if we choose learning rate $\gamma$ to be small enough $\gamma<\frac{2}{L(1+c^2)}$, we have that $\gamma-\frac{1+c^2}{2}L\gamma^2 >0$ and thus $\mathcal{L}(w^{t+1}) < \mathcal{L}(w^t)$, unless  the gradient has zero norm. Eq. \eqref{eqS32} shows us applying update rule of \methodname{} can reach the optimal value $\mathcal{L}(w^*)$ since the objective function strictly decreases.

\end{document}